\pdfoutput=1
\documentclass{article}

\PassOptionsToPackage{numbers}{natbib}
\usepackage[main, final]{neurips_2026}

\usepackage[utf8]{inputenc}
\usepackage[T1]{fontenc}
\usepackage{hyperref}
\usepackage{url}
\usepackage{booktabs}
\usepackage{amsfonts}
\usepackage{nicefrac}
\usepackage{microtype}
\usepackage[table]{xcolor}
\usepackage{graphicx}
\usepackage{tikz}
\usetikzlibrary{arrows.meta, calc}
\usepackage{pgfplots}
\pgfplotsset{compat=1.18}
\usepackage{pifont}
\usepackage{multirow}
\usepackage{caption}
\usepackage{subcaption}
\usepackage{amsthm}

\definecolor{shvpcolor}{HTML}{f5ebe0}
\definecolor{ahvpcolor}{HTML}{edede9}

\newcommand{\cmark}{\textcolor{green!60!black}{\ding{51}}}
\newcommand{\xmark}{\textcolor{red!70!black}{\ding{55}}}

\title{Hierarchical Variational Policies for \\ Reward-Guided Diffusion}

\author{%
  Kushagra Pandey\thanks{Equal contribution.}\quad
  Farrin Marouf Sofian\footnotemark[1]\quad
  Jan Niklas Groeneveld \\
  \bfseries Felix Draxler\quad
  Stephan Mandt \\[0.6em]
  {\normalfont Department of Computer Science} \\
  {\normalfont University of California, Irvine}
  \\[0.4em]
  {\normalfont\small
    \texttt{\{pandeyk1, fmaroufs, jgroenev, fdraxler, mandt\}@uci.edu}}
}

\usepackage{amsmath,amsfonts,bm}

\def\eqref#1{equation~\ref{#1}}

\def\1{\bm{1}}

\def\rvu{{\mathbf{i}}}

\def\rvu{{\mathbf{u}}}

\def\rvx{{\mathbf{x}}}
\def\rvy{{\mathbf{y}}}

\def\vzero{{\bm{0}}}

\def\vmu{{\bm{\mu}}}

\def\vpi{{\bm{\pi}}}

\def\vx{{\bm{x}}}
\def\vy{{\bm{y}}}

\def\mI{{\bm{I}}}

\def\mM{{\bm{M}}}

\def\mS{{\bm{S}}}

\DeclareMathAlphabet{\mathsfit}{\encodingdefault}{\sfdefault}{m}{sl}
\SetMathAlphabet{\mathsfit}{bold}{\encodingdefault}{\sfdefault}{bx}{n}

\def\gA{{\mathcal{A}}}

\def\gH{{\mathcal{H}}}

\def\gL{{\mathcal{L}}}

\def\gN{{\mathcal{N}}}

\newcommand{\E}{\mathbb{E}}

\newcommand{\kl}[2]{D_{\mathrm{KL}}\!\left(#1 ~ \| ~ #2\right)}

\DeclareMathOperator*{\argmax}{arg\,max}

\usepackage{cleveref}
\newcommand{\figTeaser}{%
\begin{figure}[!t]
    \centering
    \includegraphics[width=1.0\linewidth]{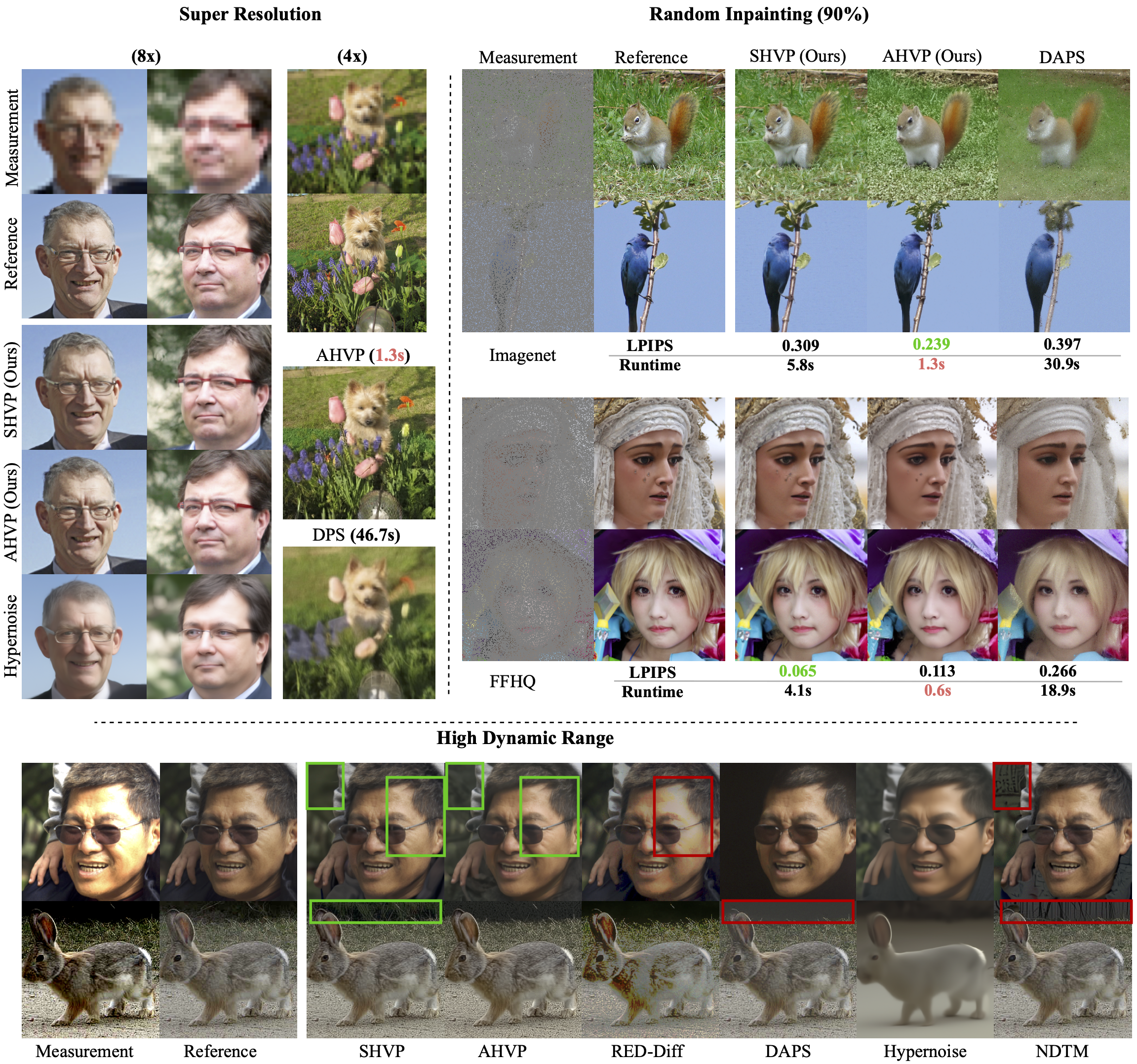}
    \caption{\textbf{AHVP and SHVP satisfy measurement constraints with high
    sample quality at reduced inference cost.} AHVP offers strong perceptual
    quality with fast inference, while SHVP typically improves quality further
    at a moderate additional cost. Shown: super-resolution ($\times8$,
    $\times4$), 90\% random inpainting, and HDR. In the HDR panel, baselines
    show artifacts (\textcolor{red}{red} boxes) that our methods avoid
    (\textcolor{green}{green} boxes). Inline numbers denote LPIPS and per-sample
    runtime. Full results in Table~\ref{tab:sr_inpainting_results}.}
    \label{fig:fig_1}
    \vspace{-0.5em}
\end{figure}%
}

\newcommand{\figAblationTwoStage}{%
\begin{figure}[!t]
    \centering
    \footnotesize
    \setlength{\tabcolsep}{0.6pt}
    \begin{tabular}{cccccc}
        & & & & \multicolumn{2}{c}{Error $|$output $-$ original$|$} \\
        \cmidrule(lr){5-6}
        Original & Degraded & Stage-1 only & Both stages & Stage-1 only & Both stages \\
        \includegraphics[width=0.163\linewidth]{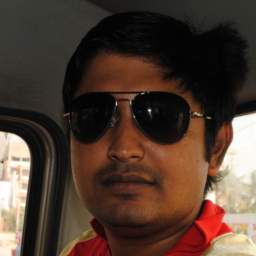} &
        \includegraphics[width=0.163\linewidth]{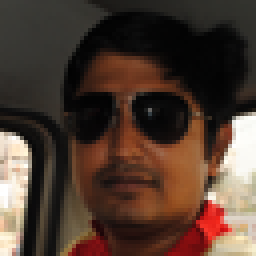} &
        \includegraphics[width=0.163\linewidth]{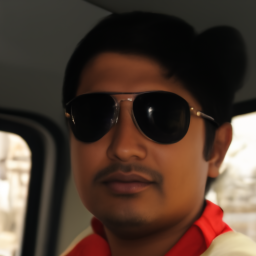} &
        \includegraphics[width=0.163\linewidth]{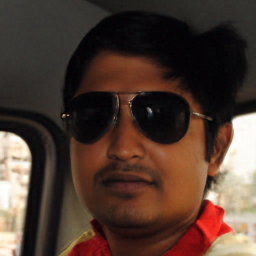} &
        \includegraphics[width=0.163\linewidth]{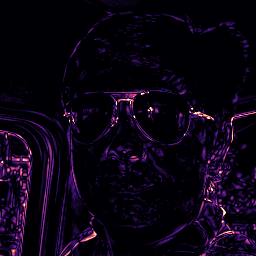} &
        \includegraphics[width=0.163\linewidth]{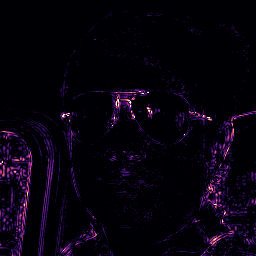} \\
        \includegraphics[width=0.163\linewidth]{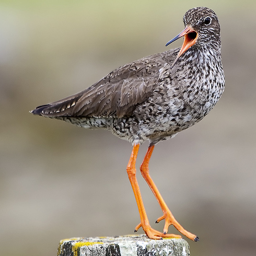} &
        \includegraphics[width=0.163\linewidth]{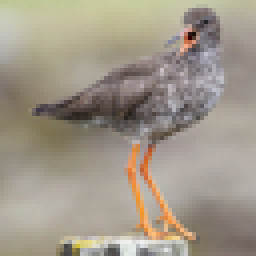} &
        \includegraphics[width=0.163\linewidth]{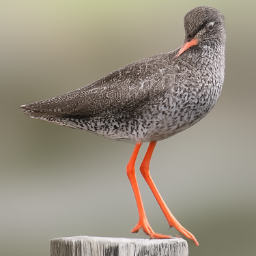} &
        \includegraphics[width=0.163\linewidth]{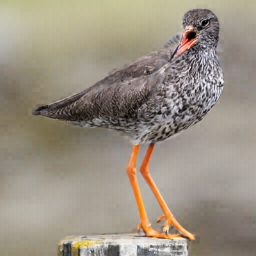} &
        \includegraphics[width=0.163\linewidth]{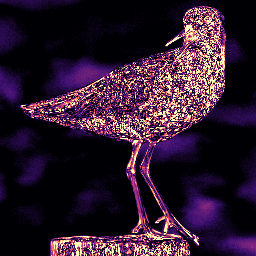} &
        \includegraphics[width=0.163\linewidth]{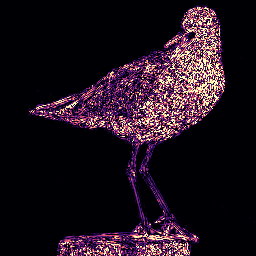} \\
        & & & & \multicolumn{2}{c}{\scriptsize $0$~\raisebox{-1pt}{\includegraphics[width=0.22\linewidth,height=5pt]{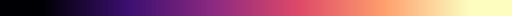}}~$0.25{+}$} \\
    \end{tabular}
    \caption{\textbf{Per-step controls recover fine details missed by the initial
    noise.} AHVP on SR$\times$4 with only the Stage-1 noise predictor vs.\ both
    stages. Error maps show $|$output $-$ original$|$ on a shared scale: Stage-1
    recovers coarse structure, and Stage-2 per-step controls further reduce the
    error. Top: FFHQ-256; bottom: ImageNet-256.}
    \label{fig:ablation_two_stage}
    \vspace{-1em}
\end{figure}%
}

\newcommand{\figAppInpainting}{%
\begin{figure}
    \centering
    \includegraphics[width=1.0\linewidth]{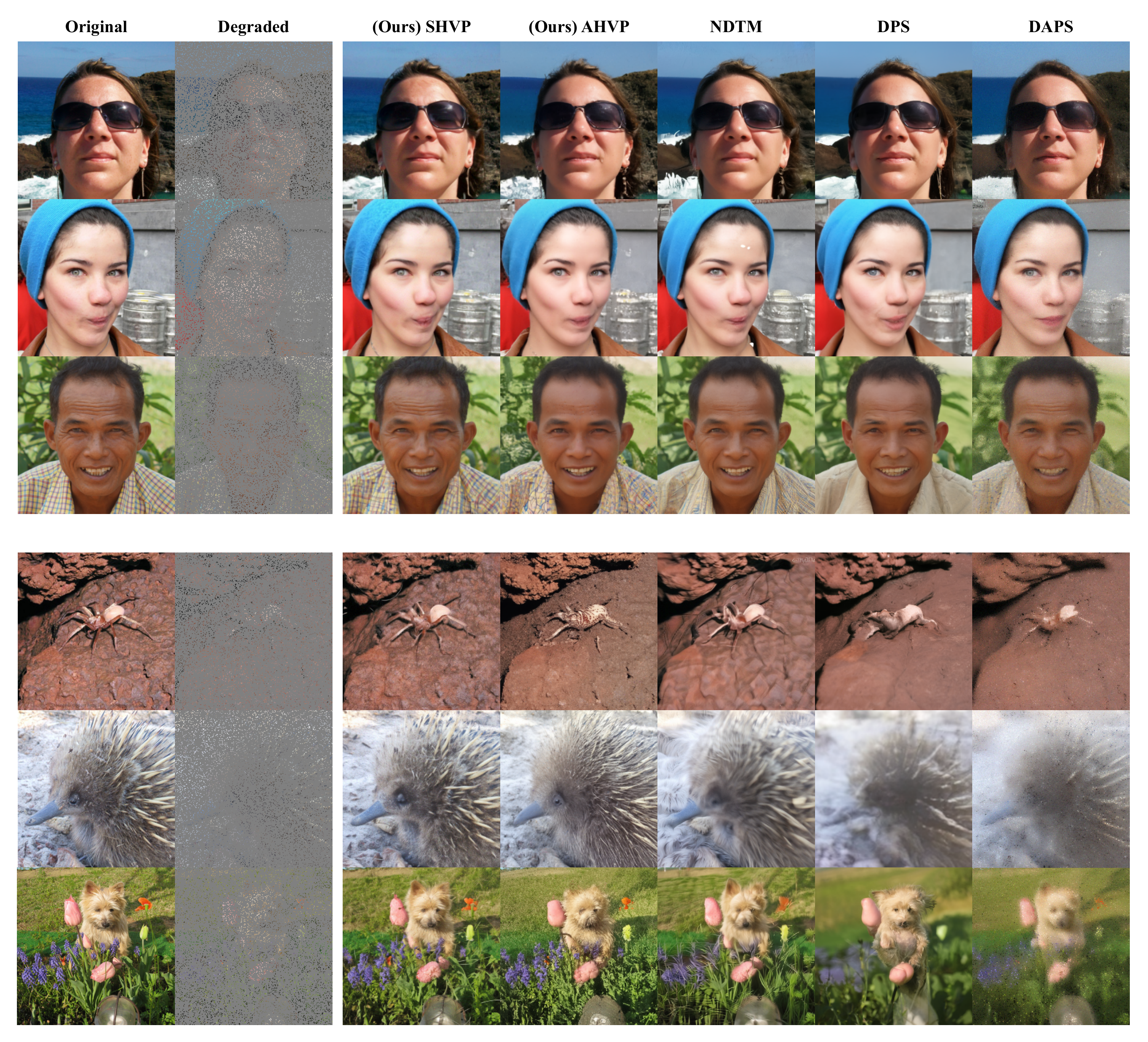}
    \caption{\textbf{Qualitative comparison on 90\% random inpainting.} Top
    panel: FFHQ-256; bottom panel: ImageNet-256. Our methods (AHVP, SHVP)
    capture fine-grained details better than competing baselines.}
    \label{fig:app_inpainting}
\end{figure}%
}

\newcommand{\figAppHDR}{%
\begin{figure}
    \centering
    \includegraphics[width=1.0\linewidth]{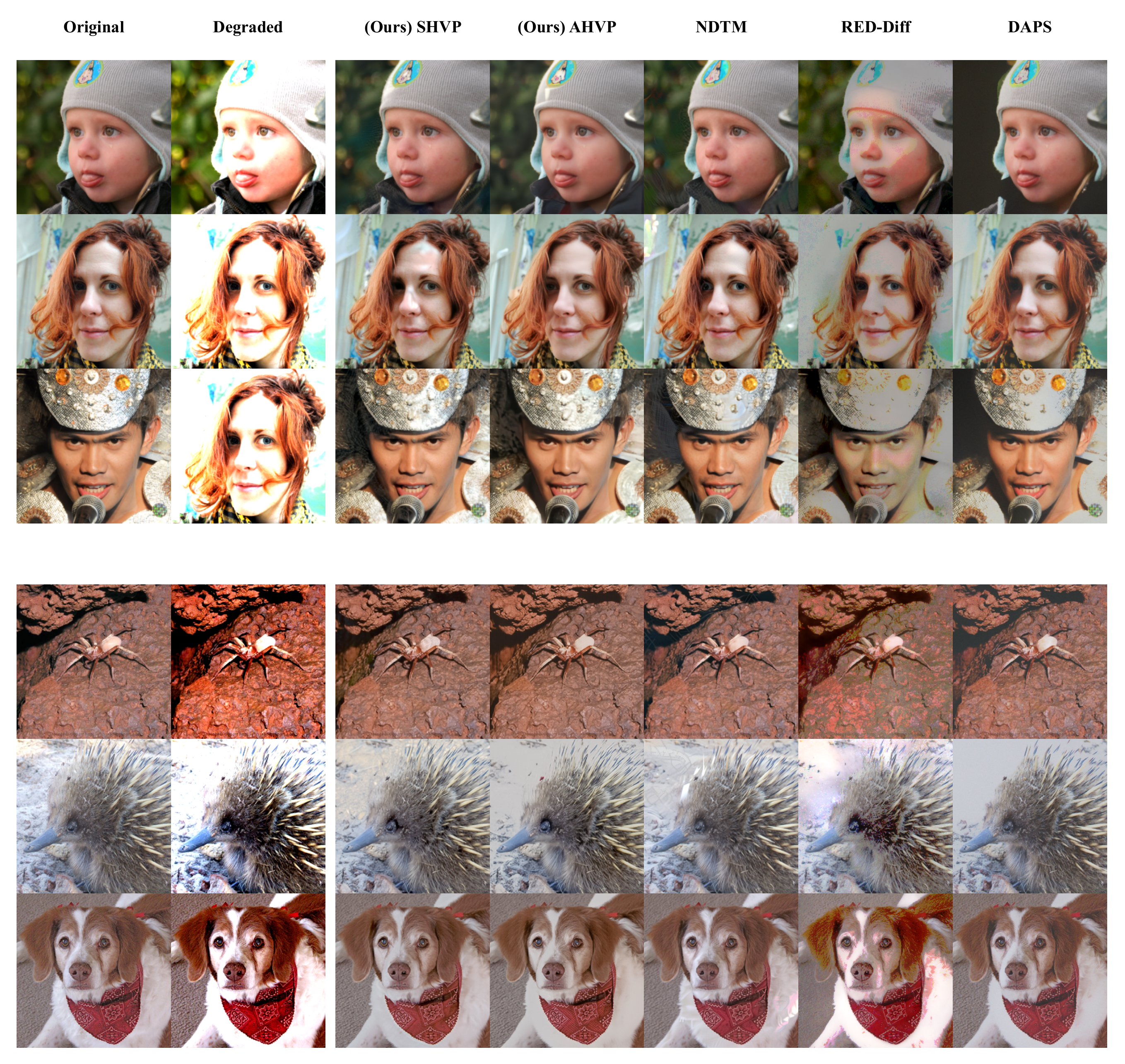}
    \caption{\textbf{Qualitative comparison on HDR.} Top panel: FFHQ-256;
    bottom panel: ImageNet-256. Our methods (AHVP, SHVP) preserve the overall
    color profile better than competing baselines.}
    \label{fig:app_hdr}
\end{figure}%
}

\newcommand{\figAppSRxEight}{%
\begin{figure}
    \centering
    \includegraphics[width=1.0\linewidth]{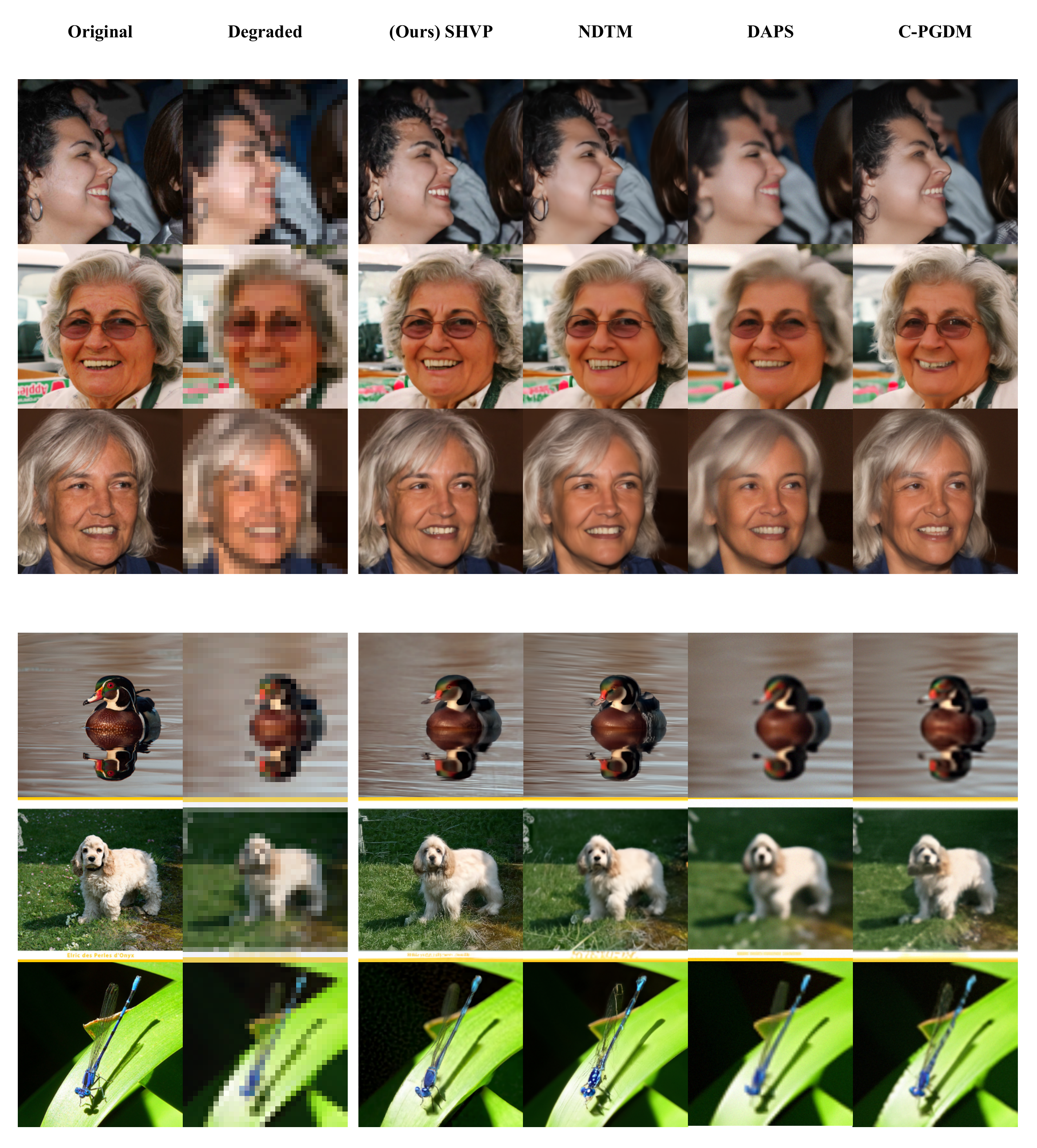}
    \caption{\textbf{Qualitative comparison on SR$\times$8.} Top panel:
    FFHQ-256; bottom panel: ImageNet-256. SHVP (ours) captures details better
    than competing baselines.}
    \label{fig:app_sr8}
\end{figure}%
}

\newcommand{\tabMainResults}{%
\begin{table*}[!t]
\centering
\footnotesize
\setlength{\tabcolsep}{5pt}
\caption{\textbf{Quantitative comparison on diffusion-based inverse problems.}
Perceptual quality (LPIPS, FID) and per-sample runtime on a single A6000 GPU;
reconstruction metrics (PSNR/SSIM) are in Table~\ref{tab:recovery}. SHVP
achieves the best perceptual quality across most tasks, while AHVP offers a
strong quality--speed tradeoff. \textbf{Bold}/\underline{underline}:
best/second-best per metric; --: unstable runs.}
\label{tab:sr_inpainting_results}
\begin{tabular}{llcccccc}
\toprule
\multirow{2}{*}{Task} & \multirow{2}{*}{Method}
& \multicolumn{3}{c}{FFHQ-256}
& \multicolumn{3}{c}{ImageNet-256} \\
\cmidrule(lr){3-5} \cmidrule(lr){6-8}
& & Time (s) $\downarrow$ & LPIPS $\downarrow$ & FID $\downarrow$
& Time (s) $\downarrow$ & LPIPS $\downarrow$ & FID $\downarrow$ \\
\midrule

\multirow{12}{*}{SR ($\times$4)}
& DAPS \citep{zhang2025improving} & 18.9 & 0.174 & 50.45 & 30.9 & 0.249 &
\underline{80.35} \\
& NDTM \citep{pandey2025variational} & 4.1 & \underline{0.069} &
\underline{41.22} & 5.8 & 0.264 & 84.15 \\
& RED-Diff \citep{reddiff} & \underline{0.3} & 0.229 & 89.67 & \underline{0.7} & 0.392
& 128.79 \\
& DDRM \citep{kawar2022denoising} & \underline{0.3} & 0.136 & 68.02 & \underline{0.7}
& 0.335 & 93.99 \\
& DPS \citep{chung2022diffusion} & 10.4 & 0.138 & 72.18 & 46.7 & 0.350 & 130.09
\\
& C-$\Pi$GDM \citep{pandey2024fast} & 0.7 & 0.111 & 63.27 & 2.4 & 0.274 & 81.44
\\
& MPGD \citep{he2024manifold} & 0.6 & 0.229 & 64.59 & 2.0 & 0.323 &
95.48 \\
& HyperNoise \citep{eyring2025noise} & 0.6 & 0.147 & 79.91 &
1.3 & 0.431 & 155.14 \\
& DEFT \citep{denker2024deft} & \textbf{0.2} & 0.148 & 63.58 & \textbf{0.5} &
0.337 & 95.76 \\
& Adjoint Matching \citep{domingo-enrich2025adjoint} & 0.6 & 0.147 & 71.84 & 1.3 &
0.349 & 127.99 \\\cmidrule(l){2-8}
\rowcolor{ahvpcolor} & (Ours) AHVP & 0.6 & 0.073 & 45.17 &
1.3 & \underline{0.244} & 84.44 \\
\rowcolor{shvpcolor} & (Ours) SHVP & 4.1 & \textbf{0.059} & \textbf{38.50} & 5.8
& \textbf{0.202} & \textbf{65.54} \\
\midrule

\multirow{12}{*}{SR ($\times$8)}
& DAPS  & 18.9 & 0.284 & 91.01 & 30.9 & 0.421 & 182.04 \\
& NDTM & 4.1 & \underline{0.145} & \underline{69.86} & 5.8 & \textbf{0.349} &
\textbf{120.85} \\
& RED-Diff  & \underline{0.3} & 0.363 & 122.74 & \underline{0.7} & 0.566 & 214.96 \\
& DDRM      & \underline{0.3} & 0.212 & 89.83 & \underline{0.7} & 0.516 & 155.31 \\
& DPS       & 10.4 & 0.192 & 81.60 & 46.7 & \underline{0.401} & 154.69 \\
& C-$\Pi$GDM   & 0.7 & 0.164 & 75.99 & 2.4 & 0.422 & 145.52 \\
& MPGD      & 0.6 & 0.395 & 100.89 & 2.0 & 0.546 & 210.43 \\
& HyperNoise& 0.6 & 0.236 & 101.72 & 1.3 & 0.606 &
212.57 \\
& DEFT & \textbf{0.2} & 0.251 & 87.81 & \textbf{0.5} & 0.462 & 167.46 \\
& Adjoint Matching & 0.6 & 0.241 & 105.67 & 1.3 & 0.557 & 189.61
\\\cmidrule(l){2-8}
\rowcolor{ahvpcolor} & (Ours) AHVP & 0.6 & 0.219 & 88.33 &
1.3 & 0.514 & 192.07 \\
\rowcolor{shvpcolor} & (Ours) SHVP & 4.1 & \textbf{0.134} & \textbf{62.48} & 5.8
& 0.445 & \underline{135.66} \\
\midrule

\multirow{12}{*}{\begin{tabular}[c]{@{}l@{}}Random \\ Inpainting \\
(90\%)\end{tabular}}
& DAPS      & 18.9 & 0.266 & 84.43 & 30.9 & 0.397 & 222.05 \\
& NDTM & 4.1 & \underline{0.107} & \underline{67.91} & 5.8 & 0.296 & 132.33 \\
& RED-Diff & \underline{0.3} & 0.827 & 305.02 & \underline{0.7} & 0.929 & 327.46 \\
& DDRM & \underline{0.3} & 0.743 & 244.08 & \underline{0.7} & 0.893 & 346.67 \\
& DPS & 10.4 & 0.122 & 69.09 & 46.7 & \underline{0.289} & 116.67 \\
& C-$\Pi$GDM       & 0.7 & -- & -- & 2.4 & -- & -- \\
& MPGD          & 0.6 & 0.354 & 118.72 & 2.0 & 0.607 & 265.72 \\
& HyperNoise & 0.6 & 0.239 & 104.60 & 1.3 & 0.484 &
188.07 \\
& DEFT & \textbf{0.2} & 0.405 & 176.05 & \textbf{0.5} & 0.549 & 258.42 \\
& Adjoint Matching & 0.6 & 0.199 & 98.28 & 1.3 & 0.403 & 210.95
\\\cmidrule(l){2-8}
\rowcolor{ahvpcolor} & (Ours) AHVP & 0.6 & 0.113 & 80.21 &
1.3 & \textbf{0.239} & \underline{111.27} \\
\rowcolor{shvpcolor} & (Ours) SHVP & 4.1 & \textbf{0.065} & \textbf{44.53} & 5.8
& 0.309 & \textbf{100.55} \\\midrule

\multirow{9}{*}{HDR}
& DAPS      & 18.9 & 0.145 & \textbf{36.22} & 30.9 & 0.186 & \underline{47.95} \\
& NDTM          & 4.1 & 0.093 & 49.51 & 5.8 & 0.142 & 55.08 \\
& RED-Diff      & \underline{0.3} & 0.138 & 72.01 & \underline{0.7} & 0.204 & 84.50 \\
& DPS           & 10.4 & -- & -- & 46.7 & -- & --\\
& HyperNoise & 0.6 & 0.193 & 102.90 & 1.3 & 0.468 &
196.98 \\
& DEFT & \textbf{0.2} & \underline{0.084} & \underline{41.00} & \textbf{0.5} &
0.142 & \textbf{44.78} \\
& Adjoint Matching & 0.6 & 0.156 & 108.50 & 1.3 & 0.203 & 96.89 \\\cmidrule(l){2-8}
\rowcolor{ahvpcolor} & (Ours) AHVP & 0.6 & 0.089 & 45.76
& 1.3 & \underline{0.138} & 61.01 \\
\rowcolor{shvpcolor} & (Ours) SHVP & 4.1 & \textbf{0.073} & 43.35 &
5.8 & \textbf{0.135} & 50.63 \\
\bottomrule
\end{tabular}
\end{table*}%
}

\newcommand{\tabAblationsSideBySide}{%
\begin{table}[t]
\centering
\begin{minipage}[b]{0.46\linewidth}
\caption{\textbf{Effect of stochastic policies.}
Impact of the policy distribution on SR$\times$4 perceptual quality
(LPIPS/FID).}
\label{tab:amortized_ablation}
\end{minipage}\hfill
\begin{minipage}[b]{0.50\linewidth}
\caption{\textbf{Per-step controls (Stage-2) improve reconstruction quality.}
Ablation on training stages for AHVP and SHVP on SR$\times$4 for FFHQ-256 and
ImageNet-256.}
\label{tab:two-stage-ablation}
\end{minipage}
\par
\begin{minipage}[t]{0.46\linewidth}
\centering
\scriptsize
\setlength{\tabcolsep}{3pt}
\begin{tabular}[t]{lcccc}
\toprule
\multirow{2}{*}{Method}
& \multicolumn{2}{c}{FFHQ-256}
& \multicolumn{2}{c}{ImageNet-256} \\
\cmidrule(lr){2-3} \cmidrule(lr){4-5}
& LPIPS $\downarrow$ & FID $\downarrow$
& LPIPS $\downarrow$ & FID $\downarrow$ \\
\midrule
AHVP (Deterministic)
& 0.081 & \textbf{43.69}
& 0.328 & 117.66 \\

AHVP (Stochastic)
& \textbf{0.073} & 45.17
& \textbf{0.244} & \textbf{84.44} \\

\bottomrule
\end{tabular}
\end{minipage}\hfill
\begin{minipage}[t]{0.50\linewidth}
\centering
\scriptsize
\setlength{\tabcolsep}{3pt}
\begin{tabular}[t]{lcccccc}
\toprule
\multirow{2}{*}{Method} & \multirow{2}{*}{Stage-1} & \multirow{2}{*}{Stage-2}
& \multicolumn{2}{c}{FFHQ-256}
& \multicolumn{2}{c}{ImageNet-256} \\
\cmidrule(lr){4-5} \cmidrule(lr){6-7}
& & & LPIPS $\downarrow$ & FID $\downarrow$
& LPIPS $\downarrow$ & FID $\downarrow$ \\
\midrule

AHVP & \cmark & \xmark
& 0.147 & 79.91
& 0.432 & 155.14 \\
AHVP & \cmark & \cmark
& \textbf{0.073} & \textbf{45.17}
& \textbf{0.244} & \textbf{84.44} \\
\midrule

SHVP & \cmark & \xmark
& 0.066 & 40.95
& 0.225 & 66.42 \\
SHVP & \cmark & \cmark
& \textbf{0.059} & \textbf{38.50}
& \textbf{0.203} & \textbf{65.54} \\

\bottomrule
\end{tabular}
\end{minipage}
\end{table}%
}

\newcommand{\tabPolicyTransfer}{%
\begin{table}[t]
\centering
\begin{minipage}[b]{0.47\linewidth}
\caption{\textbf{Same-degradation transfer.} LPIPS$\downarrow$ on FFHQ-256
of policies trained on one task (rows) and evaluated on another
(columns).}
\label{tab:transfer_same}
\end{minipage}\hfill
\begin{minipage}[b]{0.47\linewidth}
\caption{\textbf{Cross-degradation transfer and multitask training.} Same
protocol as Table~\ref{tab:transfer_same}; the last row is a single
task-conditioned controller trained jointly on both tasks.}
\label{tab:transfer_cross}
\end{minipage}
\par
\begin{minipage}[t]{0.47\linewidth}
\centering
\scriptsize
\setlength{\tabcolsep}{6pt}
\begin{tabular}[t]{lcc}
\toprule
& \multicolumn{2}{c}{Evaluated on} \\
\cmidrule(lr){2-3}
Trained on & SR$\times$4 & SR$\times$8 \\
\midrule
SR$\times$4 & \textbf{0.073} & 0.243 \\
SR$\times$8 & 0.114 & \textbf{0.219} \\
\bottomrule
\end{tabular}
\end{minipage}\hfill
\begin{minipage}[t]{0.47\linewidth}
\centering
\scriptsize
\setlength{\tabcolsep}{6pt}
\begin{tabular}[t]{lcc}
\toprule
& \multicolumn{2}{c}{Evaluated on} \\
\cmidrule(lr){2-3}
Trained on & SR$\times$4 & HDR \\
\midrule
SR$\times$4 & \textbf{0.073} & 0.204 \\
HDR & 0.189 & \textbf{0.089} \\
SR$\times$4 + HDR (joint) & \underline{0.094} & \underline{0.097} \\
\bottomrule
\end{tabular}
\end{minipage}
\end{table}%
}

\newcommand{\tabKappaAndTrainingCost}{%
\begin{table}[t]
\centering
\begin{minipage}[b]{0.48\linewidth}
\caption{\textbf{Sensitivity to $\kappa$} (policy width) on SR$\times$4
(FFHQ-256). LPIPS lower is better.}
\label{tab:kappa}
\end{minipage}\hfill
\begin{minipage}[b]{0.48\linewidth}
\caption{\textbf{Controller training cost vs.\ quality} on SR$\times$4
(FFHQ-256). LPIPS lower is better.}
\label{tab:training_cost}
\end{minipage}

\begin{minipage}[t]{0.48\linewidth}
\centering
\scriptsize
\setlength{\tabcolsep}{3pt}
\begin{tabular}[t]{lcccc}
\toprule
Inference $\kappa$ & 0.01 & 0.05 & 0.1 & 0.2 \\
\midrule
Train $\kappa =$ inference $\kappa$ & 0.083 & 0.084 & 0.079 & 0.077 \\
Train $\kappa = 0.05$ & 0.085 & 0.084 & 0.079 & 0.091 \\
\bottomrule
\end{tabular}
\end{minipage}\hfill
\begin{minipage}[t]{0.48\linewidth}
\centering
\scriptsize
\setlength{\tabcolsep}{3pt}
\begin{tabular}[t]{lccccc}
\toprule
Training steps & 1 & 100 & 500 & 1000 & 1700 \\
\midrule
LPIPS $\downarrow$ & 0.150 & 0.094 & 0.102 & 0.080 & 0.076 \\
Time (h) & 0.01 & 0.45 & 2.22 & 4.44 & 7.57 \\
\bottomrule
\end{tabular}
\end{minipage}
\end{table}%
}

\newcommand{\tabCPiGDMHparams}{%
\begin{table}[!t]
\centering
\small
\caption{Hyperparameters of C-$\Pi$GDM used for different super-resolution
tasks.}
\label{table:cpigdm_hparams}
\setlength{\tabcolsep}{6pt}
\begin{tabular}{lcccccc}
\toprule
& \multicolumn{3}{c}{FFHQ} & \multicolumn{3}{c}{ImageNet} \\
\cmidrule(lr){2-4} \cmidrule(lr){5-7}
Task & $\lambda$ & $w$ & $\tau$ & $\alpha$ & $w$ & $\tau$ \\
\midrule
SR ($\times4$) & $-0.3$ & $4.0$ & $0.4$ & $-0.4$ & $4.0$ & $0.4$ \\
SR ($\times8$) & $-0.2$ & $4.0$ & $0.4$ & $-0.5$ & $4.0$ & $0.4$ \\
\bottomrule
\end{tabular}
\end{table}%
}

\newcommand{\tabREDDiffHparams}{%
\begin{table}[t]
\centering
\small
\caption{RED-Diff hyperparameters used for different tasks.}
\label{table:reddiff_hparams}
\setlength{\tabcolsep}{6pt}
\begin{tabular}{lcccc}
\toprule
& \multicolumn{2}{c}{FFHQ} & \multicolumn{2}{c}{ImageNet} \\
\cmidrule(lr){2-3} \cmidrule(lr){4-5}
Task & lr & $\lambda$ & lr & $\lambda$ \\
\midrule
SR ($\times4$) & 0.2 & 0.1 & 0.2 & 0.1 \\
SR ($\times8$) & 0.1 & 0.1 & 0.2 & 0.1 \\
Inpainting     & 0.2 & 0.1 & 0.2 & 0.1 \\
HDR            & 0.5 & 0.1 & 0.2 & 0.5 \\
\bottomrule
\end{tabular}
\end{table}%
}

\newcommand{\tabNDTMHparams}{%
\begin{table}[!t]
\centering
\small
\caption{NDTM hyperparameters used for different tasks.}
\label{table:ndtm_hparams}
\setlength{\tabcolsep}{6pt}
\begin{tabular}{lcccccccccc}
\toprule
& \multicolumn{5}{c}{FFHQ} & \multicolumn{5}{c}{ImageNet} \\
\cmidrule(lr){2-6} \cmidrule(lr){7-11}
Task & $N$ & $\gamma$ & $\eta$ & $\tau$ & lr & $N$ & $\gamma$ & $\eta$ & $\tau$
& lr \\
\midrule
SR ($\times4$) & 5 & 1.0 & 0.0 & 400 & 0.05 & 2 & 2.0 & 0.1 & 600 & 0.05 \\
SR ($\times8$) & 5 & 1.0 & 0.0 & 500 & 0.05 & 2 & 1.0 & 0.0 & 600 & 0.07 \\
Inpainting     & 2 & 6.0 & 0.0 & 500 & 0.05 & 2 & 1.0 & 0.0 & 600 & 0.1 \\
HDR            & 5 & 1.0 & 0.2 & 400 & 0.05 & 5 & 1.0 & 0.0 & 400 & 0.05 \\
\bottomrule
\end{tabular}
\end{table}%
}

\newcommand{\tabStageOneLossWeights}{%
\begin{table}[!t]
\centering
\small
\caption{Loss weighting for Stage-1 training.}
\label{table:stage_1_hyperparams}
\setlength{\tabcolsep}{6pt}
\begin{tabular}{lcccc}
\toprule
& \multicolumn{2}{c}{FFHQ} & \multicolumn{2}{c}{ImageNet} \\
\cmidrule(lr){2-3} \cmidrule(lr){4-5}
Task & $w_T$ & $w_{\text{control}}$ & $w_T$ & $w_{\text{control}}$ \\
\midrule
SR ($\times4$) & 50.0 & 1.0 & 50.0 & 1.0 \\
SR ($\times8$) & 50.0 & 1.0 & 50.0 & 1.0 \\
Inpainting     & 50.0 & 1.0 & 50.0 & 1.0 \\
HDR            & 10.0 & 1.0 & 1.0 & 1.0 \\
\bottomrule
\end{tabular}
\end{table}%
}

\newcommand{\tabControllerArch}{%
\begin{table}[!t]
\centering
\small
\setlength{\tabcolsep}{6pt}
\caption{Architecture details of the latent and per-step controllers.}
\label{tab:controller_arch}
\begin{tabular}{lcccc}
\toprule
& \multicolumn{2}{c}{Latent Controller} & \multicolumn{2}{c}{Per-step
Controller} \\
\cmidrule(lr){2-3} \cmidrule(lr){4-5}
& FFHQ & ImageNet & FFHQ & ImageNet \\
\midrule
In-Channels    & 6   & 6   & 9   & 9   \\
Out-Channels   & 3   & 3   & 3   & 3   \\
Embed Dim.     & 768 & 768 & 768 & 768 \\
Num Heads      & 8   & 8   & 8   & 8   \\
Patch Size     & 4   & 4   & 4   & 4   \\
Num DiT Blocks & 4   & 8   & 4   & 8   \\
Dropout        & 0.0 & 0.0 & 0.0 & 0.0 \\
MLP Ratio      & 4.0 & 4.0 & 4.0 & 4.0 \\
\midrule
Parameters     & 51M & 92M & 51M & 92M \\
\bottomrule
\end{tabular}
\end{table}%
}

\newcommand{\tabStageTwoLossWeights}{%
\begin{table}[!t]
\centering
\small
\caption{Loss weighting for Stage-2 training.}
\label{table:stage_2_hyperparams}
\setlength{\tabcolsep}{6pt}
\begin{tabular}{lcccccc}
\toprule
& \multicolumn{3}{c}{FFHQ} & \multicolumn{3}{c}{ImageNet} \\
\cmidrule(lr){2-4} \cmidrule(lr){5-7}
Task & $w_T$ & $w_{\text{control}}$ & $w_{\text{score}}$ & $w_T$ &
$w_{\text{control}}$ & $w_{\text{score}}$ \\
\midrule
SR ($\times4$) & 50.0 & \texttt{ddim} & \texttt{ddim} & 50 & \texttt{ddim} &
\texttt{ddim} \\
SR ($\times8$) & 50.0 & \texttt{ddim} & \texttt{ddim} & 50.0 & \texttt{ddim} &
\texttt{ddim} \\
Inpainting & 25.0 & \texttt{ddim} & \texttt{ddim} & 25.0 & \texttt{ddim} &
\texttt{ddim} \\
HDR & 1.0 & \texttt{ddim} & \texttt{ddim} & 1.0 & \texttt{ddim} & \texttt{ddim}
\\
\bottomrule
\end{tabular}
\end{table}%
}

\newcommand{\tabReconstructionMetrics}{%
\begin{table*}[!t]
\centering
\small
\setlength{\tabcolsep}{6pt}
\caption{Reconstruction metrics for different inverse problems. -: unstable
runs.}
\label{tab:recovery}

\begin{tabular}{llcccc}
\toprule
\multirow{2}{*}{Task} & \multirow{2}{*}{Method}
& \multicolumn{2}{c}{FFHQ-256}
& \multicolumn{2}{c}{ImageNet-256} \\
\cmidrule(lr){3-4} \cmidrule(lr){5-6}
& & PSNR $\uparrow$ & SSIM $\uparrow$
& PSNR $\uparrow$ & SSIM $\uparrow$ \\
\midrule

\multirow{12}{*}{SR ($\times$4)}
& DAPS-200 \citep{zhang2025improving} & \textbf{30.19} & 0.794 & \textbf{25.30}
& 0.671 \\
& NDTM \citep{pandey2025variational} & 29.73 & \underline{0.849} & 22.88 & 0.676 \\
& RED-Diff \citep{reddiff} & 26.43 & 0.720 & 23.04 & 0.672 \\
& DDRM \citep{kawar2022denoising} & 29.49 & 0.834 & \underline{23.94} &
\underline{0.692} \\
& DPS \citep{chung2022diffusion} & 26.89 & 0.748 & 22.06 & 0.587 \\
& C-$\Pi$GDM \citep{pandey2024fast} & 28.10 & 0.765 & 23.30 & 0.630 \\
& MPGD \citep{he2024manifold} & 27.16 & 0.791 & 23.62 & 0.656 \\
& HyperNoise \citep{eyring2025noise} & 27.86 & 0.780 & 21.60 & 0.571 \\
& DEFT \citep{denker2024deft} & 28.96 & 0.815 & 23.77 & \textbf{0.710} \\
& Adjoint Matching \citep{domingo-enrich2025adjoint} & 29.76 & \textbf{0.851} &
23.37 & 0.648 \\\cmidrule(l){2-6}
& (Ours) AHVP & 29.33 & 0.836 & 23.38 & 0.665 \\
& (Ours) SHVP & \underline{29.90} & 0.842 & 23.67 & 0.648 \\

\midrule

\multirow{12}{*}{SR ($\times$8)}
& DAPS-200 & \textbf{26.32} & 0.691 & \textbf{22.29} & 0.523 \\
& NDTM & 25.35 & \underline{0.708} & 20.35 & 0.514 \\
& RED-Diff & 25.16 & 0.696 & 20.59 & 0.513 \\
& DDRM & \underline{25.78} & \textbf{0.726} & \underline{21.27} & \underline{0.550}
\\
& DPS & 23.36 & 0.628 & 19.69 & 0.467 \\
& C-$\Pi$GDM & 25.09 & 0.674 & 20.90 & 0.487 \\
& MPGD & 21.01 & 0.561 & 18.35 & 0.392 \\
& HyperNoise & 23.60 & 0.649 & 19.22 & 0.457 \\
& DEFT & 24.83 & 0.669 & 20.74 & 0.522 \\
& Adjoint Matching & 24.57 & 0.675 & 21.20 & \textbf{0.551} \\\cmidrule(l){2-6}
& (Ours) AHVP & 25.09 & 0.697 & 20.75 & 0.371 \\
& (Ours) SHVP & 24.95 & 0.677 & 21.16 & 0.526 \\

\midrule

\multirow{12}{*}{\begin{tabular}[c]{@{}l@{}}Random \\ Inpainting \\
(90\%)\end{tabular}}
& DAPS-200 & 25.72 & 0.707 & \underline{21.62} & 0.539 \\
& NDTM & \underline{27.21} & \underline{0.822} & 20.55 & \textbf{0.614} \\
& RED-Diff & 13.24 & 0.218 & 13.20 & 0.197 \\
& DDRM & 15.41 & 0.263 & 14.61 & 0.187 \\
& DPS & 26.94 & 0.776 & \textbf{21.74} & \underline{0.608} \\
& C-$\Pi$GDM & - & - & - & - \\
& MPGD & 21.34 & 0.632 & 14.97 & 0.268 \\
& HyperNoise & 23.26 & 0.648 & 19.55 & 0.466 \\
& DEFT & 16.20 & 0.421 & 15.35 & 0.310 \\
& Adjoint Matching & 24.77 & 0.713 & 20.33 & 0.500 \\\cmidrule(l){2-6}
& (Ours) AHVP & 26.32 & 0.788 & 20.74 & 0.598 \\
& (Ours) SHVP & \textbf{28.56} & \textbf{0.835} & 18.96 & 0.531 \\

\midrule

\multirow{9}{*}{HDR}
& DAPS-200 & \textbf{27.15} & 0.869 & \textbf{24.65} & 0.813 \\
& NDTM & 25.42 & 0.874 & 22.40 & \underline{0.828} \\
& RED-Diff & 24.80 & 0.803 & 20.57 & 0.689 \\
& DPS & - & - & - & - \\
& HyperNoise & 22.92 & 0.774 & 19.33 & 0.521 \\
& DEFT & 24.72 & \textbf{0.906} & 19.76 & \textbf{0.842} \\
& Adjoint Matching & 24.88 & 0.778 & 20.12 & 0.750 \\\cmidrule(l){2-6}
& (Ours) AHVP & 24.30 & 0.893 & 21.93 & 0.799 \\
& (Ours) SHVP & \underline{25.52} & \underline{0.902} & \underline{22.65} & 0.804
\\

\bottomrule
\end{tabular}
\end{table*}%
}

\begin{document}

\maketitle

\begin{abstract}
Adapting pretrained diffusion models to downstream objectives such as inverse
problems often requires expensive test-time guidance or optimization. We propose
a principled framework for generating high-quality reward-aligned samples at
substantially reduced inference cost. Our approach formulates test-time
adaptation as a hierarchical variational model, where control is amortized into
a lightweight yet expressive stochastic policy. This formulation naturally
supports few-step diffusion sampling: large step sizes enable fast inference,
while the learned policy maintains sample quality by providing structured
per-step control. The resulting fully amortized sampler achieves a strong
quality--speed tradeoff, matching or exceeding recent test-time scaling
baselines while requiring significantly less compute. For example, on $4\times$
super-resolution, our method achieves better perceptual quality with more than
\textbf{5$\times$ faster} inference compared to the best-performing baseline. We
further extend our approach to a semi-amortized regime that combines cheap
amortized proposals with limited test-time optimization, achieving
state-of-the-art perceptual quality across several challenging inverse problems.
Code and model checkpoints will be made available at
\url{https://github.com/mandt-lab/hvp}.
\end{abstract}

\section{Introduction}
\figTeaser

Diffusion models \citep{ho2020denoising, sohl2015deep} and related families
\citep{albergo2023stochastic, lipman2023flow} are powerful priors for downstream
generative tasks. Test-time adaptation of diffusion models steers the denoising
process to solve novel tasks such as text-to-image alignment
\citep{ma2025inferencetimescalingdiffusionmodels, singhal2025a,
uehara2025inferencetimealignmentdiffusionmodels} and inverse problems
\citep{chung2022diffusion, reddiff, zhang2025improving, pandey2025variational,
zheng2025inversebench, geyfman2026calibratedtesttimeguidancebayesian} without
expensive training of conditional models: just specify the task, e.g., via
a likelihood term or a learned reward \citep{xu2023imagereward, wu2023human},
and sample. However, most existing methods rely on repeated gradient evaluations
or inner-loop optimizations that can be prohibitive for
high-resolution or real-time scenarios.

A natural remedy is to amortize this cost: train a lightweight policy that
\textit{learns to steer} denoising, shifting expense from inference to a
one-time training phase. However, such policies are difficult to learn,
especially for the multimodal posteriors arising in ill-posed inverse problems
or large-step diffusion \citep{salimans2022progressive, song2023consistency,
yin2024one, geng2025mean}. We address this by adding latent variables to the
policy, yielding a more flexible hierarchical variational model
\citep{pmlr-v48-ranganath16}. We train these policies through a variational
inference formulation \citep{Blei_2017,zhang2018advances}: given a pretrained
prior and a downstream reward, we approximate the posterior over denoising
trajectories. Several recent methods
\citep{geyfman2026calibratedtesttimeguidancebayesian, pandey2025variational,
reddiff} adopt a similar view but still require expensive optimization at every
inference step. Our learned policy replaces that cost with a single forward
pass, while also jointly optimizing control signals across all sampling steps.

Our framework provides a general and modular procedure for constructing
amortized variational policies for a wide range of downstream tasks with
differentiable likelihoods or rewards.
As a concrete instantiation, we propose a two-stage procedure: first learn an
initial noise distribution that maximizes the reward, then train per-step
stochastic controllers that capture residual structure. We further extend the
framework to a semi-amortized regime that blends cheap amortized proposals with
additional test-time refinement. Empirically, on challenging inverse problems,
our approach matches or exceeds state-of-the-art test-time scaling baselines
while significantly reducing inference cost.

\begin{itemize}
    \item We introduce a \textbf{unified framework} that recasts test-time
    guidance in diffusion models as inference over hierarchical variational
    policies, enabling principled amortized control across a broad class of
    downstream tasks with differentiable rewards.

    \item We develop \textbf{Amortized HVP (AHVP)}, a novel two-stage method
    that jointly learns an initial noise distribution and per-step stochastic
    policies, producing high-quality reward-aligned samples in a single forward
    rollout.

    \item We design \textbf{Semi-Amortized HVP (SHVP)}, which combines amortized
    proposals with lightweight test-time refinement and \textbf{achieves
    state-of-the-art perceptual quality} on several challenging inverse problems
    at modest additional cost.

    \item We demonstrate a \textbf{superior quality--speed} tradeoff across four
    inverse problems on FFHQ-256 and ImageNet-256, with AHVP matching or
    exceeding the perceptual quality of leading test-time methods at
    \textbf{more than 5$\times$ faster inference} (Figs.~\ref{fig:fig_1}
    and~\ref{fig:speed_quality_tradeoff}).
\end{itemize}

\begin{figure}[t]
\centering
\input{tex/assets/speed_quality_tradeoff.tex}%

\begin{subfigure}{0.49\linewidth}
    \centering
    \plotTradeoffFFHQ
    \caption{SR ($\times4$) FFHQ-256}
\end{subfigure}
\hfill
\begin{subfigure}{0.49\linewidth}
    \centering
    \plotTradeoffImageNet
    \caption{SR ($\times4$) ImageNet-256}
\end{subfigure}

\caption{\textbf{AHVP and SHVP achieve a superior quality--speed tradeoff.}
LPIPS (lower is better) versus runtime per sample (log scale) for
diffusion-based solvers on $\times4$ super-resolution.}
\label{fig:speed_quality_tradeoff}

\end{figure}%

\section{Hierarchical Variational Policies for Reward Guidance}
Intuitively, our method amortizes guidance by learning how to nudge an
unconditional diffusion trajectory.
At each denoising step, a lightweight controller observes the current state,
timestep, and condition $\vy$, predicts a control $\rvu_t$, and the pretrained
denoiser is applied to the controlled state.
The diffusion model remains fixed; task adaptation is carried by the learned
initial-noise predictor and per-step controller.
Together, the fixed denoiser and controller act like a task-conditional sampler,
with the controller injecting information from $\vy$ while leaving the
pretrained diffusion model unchanged.
Inference therefore becomes a simple conditional rollout rather than an
inner-loop optimization.
We train the per-step controller with a reward or likelihood objective using a
Tweedie-estimate surrogate.
Below we formalize this procedure as variational inference over controlled
denoising trajectories: first as \emph{hierarchical variational policies} (HVP)
(\cref{sec:2_1}), then through a tractable learning objective (\cref{sec:2_2}),
and finally through amortized parameterizations of the resulting policies
(\cref{sec:2_3}).
\subsection{Guidance with Hierarchical Variational Policies (HVP)}
\label{sec:2_1}

We now present guidance in diffusion models through the lens of hierarchical
variational approximations. Given an observed conditioning signal $\vy$, we
define a generative process that gradually converts noise into samples
conditioned on $\vy$. More specifically,
\begin{equation}
    p(\rvx_{0:T}, \vy) = p(\rvx_T)\left[\prod_{t=1}^{T} p(\rvx_{t-1} \mid
    \rvx_t)\right]p(\vy\mid\rvx_0),
    \label{eq:p_process_main}
\end{equation}
where $p(\rvx_T)$ denotes the prior over the initial noise (commonly a standard
Gaussian distribution) and $p(\rvx_{t-1} \mid \rvx_t)$ intuitively infers a
denoised state $\rvx_{t-1}$, given the previous noisy state $\rvx_t$
\citep{ho2020denoising, albergo2023stochastic, lipman2023flow}. The distribution
$p(\vy\mid\rvx_0)$ represents the likelihood of observing $\vy$ given the final
denoised state $\rvx_0$. In practice, the form of the likelihood is
application-specific. For instance, in inverse problems (like inpainting), it
can be defined as $p(\vy\mid\rvx_0) = \gN(\vy; \gA(\rvx_0), \sigma_y^2 \mI_d)$
where $\gA$ \emph{corrupts} the denoised sample $\rvx_0$. In this work, we
assume that the likelihood distribution is \textbf{known} and
\textbf{differentiable}.

Our goal is to infer the intermediate states $\rvx_{0:T}$ that maximize the
likelihood of $\vy$. Although the main object of interest is $p(\rvx_0 \mid
\vy)$, approximating the full trajectory posterior $p(\rvx_{0:T} \mid \vy)$
allows us to reuse the diffusion generative dynamics for downstream tasks. Since
this posterior is intractable, we introduce a variational distribution
$q(\rvx_{0:T} \mid \vy)$. The resulting Evidence Lower Bound (ELBO) is
\begin{align}
    \log p(\vy) &= \log \int q(\rvx_{0:T}\mid \vy) \frac{p(\rvx_{0:T},
    \vy)}{q(\rvx_{0:T}\mid \vy)} d\rvx_{0:T} \;\;
    \geq \E_{q(\rvx_{0:T} \mid \vy)}\left[\log p(\rvx_{0:T}, \vy) - \log
    q(\rvx_{0:T}\mid\rvy) \right].
    \label{eq:elbo_initial}
\end{align}
The variational distribution $q(\rvx_{0:T}\mid\vy)$ can be interpreted as a
\emph{stochastic policy} over diffusion trajectories. While
Eq.~\ref{eq:p_process_main} defines the unguided diffusion dynamics, optimizing
the ELBO in Eq.~\ref{eq:elbo_initial} amounts to learning a controlled process
that steers these dynamics toward states consistent with $\vy$.
To optimize the objective in \cref{eq:elbo_initial}, we factorize the
variational distribution as
\begin{equation}
    q(\rvx_{0:T}\mid\vy) = q(\rvx_T\mid\vy)\prod_{t=1}^{T} q(\rvx_{t-1}
    \mid\rvx_t, \vy).
\end{equation}
We now need to choose a parameterization for each $q(\rvx_{t-1}\mid\rvx_t,
\vy)$. We find that simple choices such as unimodal Gaussians are not
sufficient:
For example, in inverse problems like inpainting where $\vy$ denotes an image
with missing pixels, multiple plausible completions may exist for $\rvx_{t-1}$
given $\rvx_t$ and $\vy$, inducing strong multimodality.
This also holds for larger denoising step sizes which are known to induce more
complex and potentially multimodal conditional denoising
distributions~\citep{song2023consistency,geng2025mean,boffi2025flow,zhou2025inductive}.
Therefore, for an expressive variational policy, we augment the variational
distribution with additional latent variables $\rvu_t$ at each time step
\citep{pmlr-v48-ranganath16},

\begin{equation}
    q(\rvx_{0:T}\mid\vy) = \int q(\rvx_{0:T}, \rvu_{1:T} \mid \vy)\;
    d\rvu_{1:T}.
    \label{eq:q_hvp}
\end{equation}
Through the lens of optimal control, these latent variables $\rvu_{1:T}$ can be
interpreted as \emph{stochastic controls}.
Owing to this hierarchical structure with marginalized latent controls, we refer
to the resulting formulation as \emph{Hierarchical Variational Policies (HVP)},
which we define as,
\begin{equation}
  q(\rvx_{0:T}, \rvu_{1:T} \mid \vy) = q(\rvx_T\mid\vy) \prod_{t=1}^T
  q(\rvu_{t}\mid \rvx_t, \rvu_{>t}, \vy)q(\rvx_{t-1}\mid\rvx_t, \rvu_t).
\end{equation}
The \emph{initial noise} distribution $q(\rvx_T\mid\vy)$ predicts the initial
noise given the side information. The \emph{per-step policy}
$q(\rvu_t\mid\rvx_t, \rvu_{>t}, \vy)$ outputs the controls at time $t$
conditioned on the current state, side information, and past controls; in
early experiments, we found that this conditioning improves training
stability and performance (see App. C). Next, we define the
\emph{conditional state transition} as,
\begin{equation}
q(\rvx_{t-1}\mid\rvx_t, \rvu_t) = p(\rvx_{t-1}\mid\rvx_t + \gamma \rvu_t),
\label{eq:guided_transition}
\end{equation}
where $p(\rvx_{t-1}\mid\rvx_t)$ is just the denoising distribution of the
generative model, and $\gamma$ is a tuning parameter. Thus, the controls
$\rvu_t$ act as additive corrections on the states $\rvx_t$ that the diffusion
model operates on~\citep{pandey2025variational}. This simple construction leaves
us with only learning the policy and initial noise distributions.
Next, we derive a tractable objective for learning such policies.
\subsection{Policy Learning via Hierarchical Variational Bounds}
\label{sec:2_2}
In its naive form, variational inference becomes intractable when the
variational distribution itself is defined as an integral over latent
variables~\citep{Blei_2017,zhang2018advances}. In this case, the entropy of the
resulting mixture distribution cannot be computed in closed form. Fortunately,
advances in variational inference over the past decade provide a tractable
alternative. Inspired by prior work~\citep{pmlr-v48-ranganath16}, we adopt a
less commonly used formulation of variational inference that derives a tractable
objective by introducing a lower bound on the entropy term. Due to the length of
the derivation, we present only the final result here; full details are in
App.~A.
The hierarchical ELBO decomposes into four interpretable terms: a reward term
and three regularizers that ensure samples are likely under the unguided
diffusion model,
\begin{align}
    \log p(\vy) &\geq \E_{\rvx_0 \sim q} \big[\log p(\vy\mid\rvx_0)\big] -
    \underbrace{\kl{q(\rvx_T\mid\vy)}{p(\rvx_T)}}_{\gL_1} \notag\\
    &\qquad \qquad -\sum_t \E_{\rvx_t, \rvu_t}
    \big[\underbrace{\kl{q(\rvx_{t-1}\mid \rvx_t, \rvu_t)}{p(\rvx_{t-1} \mid
    \rvx_t)}}_{\gL_2}\big] \notag\\
    &\qquad \qquad -\sum_t \E_{\rvx_t,
    \rvu_{>t}}\big[\underbrace{\kl{q(\rvu_t\mid\rvx_t, \rvu_{>t},
    \vy)}{\bar{r}(\rvu_t)}}_{\gL_3}\big].%
    \label{eq:elbo_final}
\end{align}
The regularizer $\bar{r}(\rvu_t)$ over the controls is a design decision. We
choose a standard Gaussian distribution $\bar{r}(\rvu_t) = \gN(\vzero, \mI_d)$.
From a controls perspective, the first term in \cref{eq:elbo_final} encourages
the guided process $q$ to generate samples that maximize the likelihood $p(\vy
\mid \rvx_0)$.
In applications where the explicit form of the likelihood is not available, we
can instead choose to maximize the expected reward $r(\rvx_0, \vy)$, which is a
measure of the alignment between the generated sample $\rvx_0$ and the
conditioning signal $\vy$ (for instance, human preference scores given a text
prompt and a generated sample
\citep{xu2023imagereward,wu2023human,kirstain2023pick}).
The second term in Eq. \ref{eq:elbo_final} regularizes the noise policy with the
unconditional noise prior $p(\rvx_T)$. The third term in Eq. \ref{eq:elbo_final}
encourages the state transition distribution to stay \emph{close} to the
unguided dynamics of the pretrained iterative refinement model, implying high
likelihood of the generated samples under the pretrained model. This regularizer
is commonly used in RL-based finetuning of diffusion models
\citep{uehara2024fine,
uehara2024understandingreinforcementlearningbasedfinetuning} and has shown to be
useful in preventing \emph{reward hacking}, i.e., overfitting to the reward
without generating plausible samples. The last term in Eq. \ref{eq:elbo_final}
encourages the per-step policy $q(\rvu_t \mid \rvx_t, \rvu_{>t}, \vy)$ to be
regularized by the prior over the controls $\bar{r}(\rvu_t)$.
Next, we discuss parameterizations of different distributions in Eq.
\ref{eq:elbo_final}.

\begin{figure}[t]
    \centering
    \definecolor{hvpInk}{HTML}{2E2E2E}
\definecolor{hvpVar}{HTML}{F2F2F2}
\definecolor{hvpVarLine}{HTML}{8C8C8C}
\definecolor{hvpObs}{HTML}{EED6E4}
\definecolor{hvpObsLine}{HTML}{A8327A}
\definecolor{hvpCtl}{HTML}{F2D8D1}
\definecolor{hvpCtlLine}{HTML}{C4502F}
\definecolor{hvpOneBg}{HTML}{EEEDF8}
\definecolor{hvpOneLine}{HTML}{4B44B0}
\definecolor{hvpTwoBg}{HTML}{FBECE6}
\definecolor{hvpTwoLine}{HTML}{C4502F}
\definecolor{hvpFrozen}{HTML}{F4F4F4}
\definecolor{hvpFrozenLine}{HTML}{8C8C8C}
\begin{tikzpicture}[
    font=\small,
    every node/.style={text=hvpInk},
    var/.style={circle, fill=hvpVar, draw=hvpVarLine, line width=0.6pt,
        minimum size=0.72cm, inner sep=1pt},
    obs/.style={var, fill=hvpObs, draw=hvpObsLine},
    ctl/.style={var, fill=hvpCtl, draw=hvpCtlLine},
    net/.style={rectangle, rounded corners=5pt, fill=white, line width=1.1pt,
        minimum width=1.85cm, minimum height=1.15cm, align=center,
        inner xsep=5pt},
    frozen/.style={net, fill=hvpFrozen, draw=hvpFrozenLine,
        dash pattern=on 3pt off 2pt, line width=0.9pt},
    plus/.style={circle, fill=white, draw=hvpInk, line width=0.7pt,
        minimum size=0.38cm, inner sep=0pt,
        path picture={
            \draw[line width=0.8pt, hvpInk]
                ([xshift=-0.1cm]path picture bounding box.center) --
                ([xshift= 0.1cm]path picture bounding box.center)
                ([yshift=-0.1cm]path picture bounding box.center) --
                ([yshift= 0.1cm]path picture bounding box.center);}},
    panel title/.style={font=\scriptsize\sffamily\bfseries, inner sep=0pt},
    note/.style={font=\scriptsize, inner sep=1pt},
    tip/.style={-{Latex[round, length=1.9mm, width=1.6mm]}},
    flow/.style={tip, line width=0.8pt, draw=hvpInk},
    cond/.style={tip, line width=0.8pt, draw=hvpObsLine!80!black},
    ctlflow/.style={tip, line width=0.9pt, draw=hvpCtlLine,
        dash pattern=on 3.5pt off 2pt},
]

\filldraw[fill=hvpOneBg, draw=hvpOneLine!35, line width=0.6pt,
    rounded corners=7pt] (0,0) rectangle (4.7,3.35);
\filldraw[fill=hvpTwoBg, draw=hvpTwoLine!35, line width=0.6pt,
    rounded corners=7pt] (6.0,0) rectangle (12.75,3.35);
\node[panel title, text=hvpOneLine, anchor=north east]
    at (4.58,3.23) {1\,$\cdot$\,Initial noise};
\node[panel title, text=hvpTwoLine, anchor=north east]
    at (11.75,3.23) {2\,$\cdot$\,Per-step control};

\node[obs] (y) at (0.6,3.98) {$\vy$};
\node[note, font=\scriptsize\itshape, text=hvpObsLine!85!black, anchor=south west]
    at ($(y.east)+(0.08,0.02)$) {observed};
\node[var]  (eps) at (0.6,0.65)  {$\epsilon$};
\node[plus] (p1)  at (2.75,0.65) {};
\node[var]  (xT)  at (4.2,0.65)  {$\rvx_T$};
\node[net, draw=hvpOneLine] (E) at (2.75,2.15)
    {Noise Predictor\\[1pt]$E_\phi$};
\node[ctl] (upast) at (5.35,2.15) {$\rvu_{>t}$};
\node[var]  (xt)  at (7.2,0.65)   {$\rvx_t$};
\node[net, draw=hvpTwoLine] (pi) at (7.2,2.15)
    {Policy\\[1pt]$\vpi_\psi$};
\node[ctl]  (ut)  at (9.05,2.15)  {$\rvu_t$};
\node[plus] (p2)  at (9.05,0.65)  {};
\node[frozen] (D) at (11.15,2.15) {Pretrained\\[1pt]Denoiser};
\node[var]  (xtm) at (11.15,0.65) {$\rvx_{t-1}$};
\node[var]  (x0)  at (13.58,0.65) {$\rvx_0$};

\node[circle, fill=white, draw=hvpFrozenLine, line width=0.7pt,
    minimum size=0.46cm, inner sep=0pt] (lock) at (D.north east) {};
\begin{scope}[shift={(lock.center)}, hvpFrozenLine]
    \draw[line width=0.9pt] (-0.055,0.0) -- (-0.055,0.04)
        arc[start angle=180, end angle=0, radius=0.055] -- (0.055,0.0);
    \fill[rounded corners=0.6pt] (-0.09,-0.115) rectangle (0.09,0.03);
\end{scope}

\draw[line width=0.8pt, draw=hvpObsLine!80!black] (y.east) -- (E.north |- y.east);
\fill[hvpObsLine!80!black] (E.north |- y.east) circle (1.4pt);
\draw[cond] (E.north |- y.east) -- (E.north);
\draw[cond] (E.north |- y.east) -- (pi.north |- y.east) -- (pi.north);

\draw[flow] (eps.north) |- (E.west);
\draw[flow] (E) -- (p1);
\draw[flow] (eps) -- node[note, above] {$\sim\gN(\vzero,\mI)$} (p1);
\draw[flow] (p1) -- (xT);

\draw[flow] (xT) -- (xt);
\node[note, above] at ($(xT.east)!0.5!(xt.west)$) {$t{=}T$};

\draw[ctlflow] (upast) -- (pi);
\draw[flow] (xt) -- (pi);
\draw[flow] (pi) -- (ut);
\draw[flow] (ut) -- (p2);
\draw[flow] (xt) -- (p2);
\draw[flow] (p2.east) -- ($(p2.east)!0.5!(D.west |- p2.east)$) |- (D.west);
\draw[flow] (D) -- (xtm);

\begin{scope}[shift={(12.75,3.35)}]
    \draw[tip, line width=1.1pt, draw=hvpTwoLine]
        (0,-0.38) arc[start angle=256.5, end angle=553.5, radius=0.514];
    \node[font=\scriptsize\bfseries, text=hvpTwoLine, inner sep=0pt]
        at (0.14,0.14) {$\times T$};
\end{scope}

\draw[flow] (xtm) -- (x0);
\node[note, above] at ($(xtm.east)!0.5!(x0.west)$) {$t{=}1$};

\end{tikzpicture}%
    \caption{\textbf{AHVP samples in a single rollout, without expensive
    test-time optimization.} (1)~The Stage-1 network $E_\phi$ predicts the
    initial noise $\rvx_T$. (2)~For $t=T,\dots,1$, the Stage-2 policy
    $\vpi_\psi$ samples a control $\rvu_t$ that perturbs $\rvx_t$, steering the
    trajectory toward samples consistent with $\vy$.}
    \label{fig:main_method}
\end{figure}%

\subsection{Amortized Hierarchical Variational Policies (AHVP)}
\label{sec:2_3}

While the distributions defined in \cref{eq:elbo_final} can be determined in a
test-time scaling setup
\citep{pandey2025variational,janati2024divideandconquer}, this is
computationally expensive during inference. Instead, we propose to amortize the
noise policy $q(\rvx_T\mid\vy)$ and the per-step policy $q(\rvu_t\mid\rvx_t,
\rvu_{>t}, \vy)$ using parametric estimators. A key advantage of amortization is
that during inference, sampling requires a simple rollout through the guided
process $q(\rvx_{0:T}, \rvu_{1:T} \mid \vy)$, therefore significantly improving
runtime over test-time scaling approaches. We choose the following
straightforward yet expressive parameterizations of these distributions:

For the \textbf{initial noise}, we implicitly model the distribution via a call
to a DiT model $E_\phi$ \cite{Peebles_2023_ICCV}:
\begin{equation}
    \rvx_T = \epsilon + E_\phi(\vy, \epsilon), \;\; \epsilon \sim \gN(0,
    \mI_d).
    \label{eq:initial_noise}
\end{equation}
This is an expressive policy, and we follow \citep{eyring2025noise} and
approximate the corresponding regularizer $\gL_1 \approx \E_{\epsilon \sim
\gN(\vzero, \mI_d)}[\Vert E_\phi(\vy, \epsilon) \Vert_2^2]$.
We model the \textbf{per-step policy} based on the pretrained diffusion model,
adding the controls $\rvu_t$ to the input of the pretrained denoising network
$\vmu_\text{pretrained}$,
\begin{equation}
    q(\rvx_{t-1} |\rvx_t, \rvu_t, \vy) = \gN(\vmu_{\text{pretrained}}(\rvx_t +
    \gamma \rvu_t, t), \sigma_{t-1}^2\mI_d).
\end{equation}
For expressivity, the \textbf{latent control variates} follow a distribution.
We choose a Gaussian,
\begin{equation}
    q(\rvu_t | \rvx_t, \rvu_{>t}, \vy) = \gN(\vpi_\psi(\rvx_t, \rvu_{>t}, \vy,
    t), \bar{\sigma}_t^2\mI_d).
    \label{eq:gauss_policy}
\end{equation}
Similar to the noise predictor, we model $\vpi_\psi(.)$ using a lightweight DiT
architecture. Since it acts on the input space to the pretrained model, it
leverages the powerful representations of the pretrained network. With this
setup, the remaining loss terms evaluate to $\gL_2 \propto \Vert \vmu(\rvx_t +
\gamma \rvu_t, t) - \vmu(\rvx_t, t) \Vert_2^2$
and
$
    \gL_3 \propto \Vert \vpi_\psi(\rvx_t, \rvu_{>t}, \vy, t) \Vert_2^2 = \|
    \rvu_t \|_2^2.
$
While the standard deviation $\bar{\sigma}_t$ could be learned, we set it
heuristically to $\bar{\sigma}_t = \kappa \sigma_t$, where $\kappa$ is a scalar
hyperparameter, and $\sigma_t$ is the standard deviation of the reverse
denoising process at time $t$ such that the noise level monotonically decreases
in the denoising process. The intuition is that the controls encode more
uncertainty during earlier denoising stages as compared to the end.

Together, this fully determines the loss function. Practically speaking, we
first train the initial noise policy network $E_\phi$ by backpropagating through
the diffusion chain. We then optimize the per-step policy $\vpi_\psi$ over all
steps in parallel by replacing the reward term with the Tweedie estimate $\log
p(\rvy \mid \hat \rvx_0)$. This avoids backpropagation through the diffusion
chain.
Experimentally, we find that learning the initial noise can recover high-level
structure, while learning the per-step policies recovers more granular details.

\textbf{Semi-Amortized Hierarchical Variational Policies (SHVP).}
Training neural networks for amortization will result in a finite amortization
error due to finite model capacity and training data. This amortization error
can be reduced by spending a small amount of additional optimization, offering a
simple way to trade off quality and speed.
We further tune the controls with a few gradient steps:
\begin{align}
    \rvu_{t}^* &= \argmax_{\rvu_{t}} \E_{q} \Big[\log p(\vy\mid\hat{\rvx}_0)
    -\lambda_2 \Vert \vmu(\rvx_t + \gamma \rvu_t, t) - \vmu(\rvx_t, t) \Vert_2^2
    - \lambda_3 \Vert \rvu_t \Vert_2^2\Big].
\end{align}
This loss follows from reducing \cref{eq:elbo_final} to the controls of the
diffusion step under consideration \citep{pandey2025variational}.
Overall, our approach follows the core idea behind semi-amortized variational
inference \citep{kim2018semi,marino2018iterative}. Therefore, we denote this
method as semi-amortized hierarchical variational policies (SHVP).

\section{Experiments}
While our framework can be applied to any differentiable reward, we demonstrate
its empirical benefits on a suite of challenging \emph{inverse problems}. Across
both quantitative and qualitative evaluations, our approach achieves a superior
perception--runtime tradeoff compared to recent state-of-the-art test-time
scaling methods. In particular, the fully amortized sampler (AHVP) provides
high-quality reconstructions with significantly reduced inference time, while
the semi-amortized variant (SHVP) consistently achieves the best perceptual
quality across tasks. We further investigate several key design choices
underlying our framework. Through ablation studies, we analyze the effect of the
policy parameterization and the impact of the proposed two-stage training
procedure. We defer implementation details
to App.~\ref{app:implementation}.

\noindent
\textbf{Reward Specification.} For inverse problems, we define the likelihood as
$p(\bm{y} \mid \mathbf{x}_0) = \mathcal{N}(\mathcal{A}(\mathbf{x}_0), \sigma_y^2
\mathbf{I}_d)$, recovering $\mathbf{x}_0 \sim p_\text{data}$ from degradations
$\bm{y} = \mathcal{A}(\mathbf{x}_0) + \sigma_y \bm{z}$, $\bm{z} \sim
\mathcal{N}(\bm{0}, \mathbf{I}_d)$. Following \citep{chung2022diffusion,
pandey2025variational}, we evaluate on linear tasks ($4\times$ and $8\times$
super-resolution; $90\%$ random inpainting) and a non-linear task (HDR with
oversaturation factor $2.0$), fixing $\sigma_y = 0.01$ across all tasks. See
App.~\ref{app:implementation} for the exact form of $\mathcal{A}$.

\textbf{Models, Datasets, and Training.} We experiment on
FFHQ~\citep{karras2018progressive} and ImageNet~\citep{deng2009imagenet} at $256
\times 256$, using publicly available unconditional pretrained
checkpoints~\citep{chung2022diffusion, dhariwal2021diffusion}. Both the noise
predictor and per-step policy use the DiT architecture, trained on 1k (FFHQ) and
10k (ImageNet) images per task. Evaluation uses a held-out set of 100 samples
per dataset. Full hyperparameters are in App.~\ref{app:implementation}.

\textbf{Baselines.} We compare AHVP and SHVP against
DAPS~\citep{zhang2025improving}, NDTM~\citep{pandey2025variational},
RED-Diff~\citep{reddiff}, C-$\Pi$GDM~\citep{pandey2024fast},
MPGD~\citep{he2024manifold}, DDRM~\citep{kawar2022denoising},
DPS~\citep{chung2022diffusion}, HyperNoise~\citep{eyring2025noise} (a
noise-only variant of Eq.~\ref{eq:elbo_final}), DEFT~\citep{denker2024deft},
and Adjoint Matching~\citep{domingo-enrich2025adjoint}. To assess test-time scaling
under limited budgets, we reduce sampling steps by 5--6$\times$ (minimum 8) and
tune all baselines for best low-step performance.

\paragraph{Evaluation Metrics.} We report perceptual metrics
(LPIPS~\citep{zhang2018unreasonable}, FID~\citep{heusel2017gans}) in the main
text and reconstruction metrics (PSNR, SSIM) in App.~\ref{app:extended_results}.

\subsection{Main Results}

Table~\ref{tab:sr_inpainting_results} presents quantitative comparisons across
four challenging inverse problems.
Across all tasks and datasets, our methods consistently achieve strong
perceptual quality. \textit{Overall, AHVP provides the best quality--runtime
tradeoff, while SHVP consistently achieves the highest perceptual quality.}

\noindent
\textbf{Perceptual Quality Comparisons.}
AHVP outperforms most test-time scaling baselines while maintaining competitive
reconstruction quality.
For example, on SR$\times4$, AHVP achieves better perceptual quality than all
test-time scaling baselines on ImageNet while remaining competitive on FFHQ.
Similarly, for random inpainting, AHVP achieves competitive performance on FFHQ
and state-of-the-art LPIPS on ImageNet.
Comparable trends are observed for the HDR task, demonstrating that AHVP
achieves a favorable perception--distortion tradeoff. To further study the role
of amortization for initializing test-time scaling methods, we evaluate
semi-amortized hierarchical policies (SHVP).
Across all four tasks, SHVP achieves the best perceptual quality in most
settings, establishing new state-of-the-art results on several benchmarks.
These results validate our hypothesis that pretrained amortized policies can
serve as strong initialization for more powerful test-time scaling strategies.
Overall, AHVP provides the best quality--runtime tradeoff,
while SHVP consistently achieves the highest perceptual quality across tasks.

\noindent

\textbf{Runtime Comparisons.}
In addition to perceptual quality, it is important to evaluate the
perception--runtime tradeoff.
Table~\ref{tab:sr_inpainting_results} reports the runtime of all competing
baselines on both FFHQ and ImageNet.
AHVP is significantly faster than most test-time scaling baselines while
maintaining strong reconstruction quality.
For instance, on the SR$\times4$ task, AHVP is more than $5\times$ faster than
NDTM while achieving similar or better perceptual quality.
This tradeoff is further illustrated in Fig.~\ref{fig:speed_quality_tradeoff}.
Although methods such as RED-Diff and DDRM provide fast inference, they produce
substantially worse perceptual quality.
Conversely, approaches such as DAPS and NDTM achieve strong quality but require
significantly more computation.
In contrast, AHVP achieves competitive perceptual quality with extremely
efficient inference, yielding a superior overall perception--runtime tradeoff.
This efficiency arises from amortization, which enables guided inference through
a single pass of the controlled process $q$ (Eq.~\ref{eq:q_hvp})
without expensive inner-loop optimization or gradient computations.
Moreover, SHVP achieves higher perceptual quality than strong baselines such as
NDTM at comparable inference cost,
demonstrating that amortized policies provide an effective foundation for
constructing efficient test-time samplers.

\tabMainResults

\subsection{Ablation Results}
\label{sec:ablations}
\figAblationTwoStage
\tabAblationsSideBySide

\noindent
\textbf{Impact of Policy Type.}
We study the effect of modeling the per-step control policy as a stochastic
distribution.
In our method, the policy is parameterized as a unimodal Gaussian
(Eq.~\ref{eq:gauss_policy}).
We compare against a deterministic policy baseline,
$q(\rvu_t \mid \rvx_t, \rvu_{>t}, \vy) = \delta(\rvu_t = \vpi_\psi(\rvx_t,
\rvu_{>t}, \vy, t))$.
Table~\ref{tab:amortized_ablation} reports the perceptual quality of stochastic
and deterministic policies on the SR$\times$4 task for FFHQ and ImageNet.
On FFHQ, both variants achieve comparable performance.
However, on ImageNet the stochastic policy substantially improves perceptual
quality,
reducing LPIPS from $0.328$ to $0.244$ and improving FID from $117.66$ to
$84.44$.
These results suggest that stochastic policies provide a more flexible control
mechanism for complex datasets.
Exploring richer policy distributions, such as normalizing-flow-based policies,
is an interesting direction for future work.

\noindent
\textbf{Impact of Two-Stage Training.}
Table~\ref{tab:two-stage-ablation} evaluates Stage-2 controls on top of the
Stage-1 noise policy for SR$\times4$.
Stage-2 reduces the LPIPS of AHVP by nearly $50\%$ and consistently improves
SHVP.
Fig.~\ref{fig:ablation_two_stage} shows that per-step controls recover sharp
edges and textures missed by the initial noise policy. Thus, Stage-1 provides a
strong trajectory initialization, while Stage-2 supplies fine-grained
refinements along the rollout.

\tabPolicyTransfer

\noindent
\textbf{Policy Transfer.}
We test whether learned policies transfer across degradations on FFHQ, both
within a degradation family (SR$\times$4 $\leftrightarrow$ SR$\times$8) and
across degradations (SR$\times$4 $\leftrightarrow$ HDR). Policies transfer
within a family: applied to SR$\times$8, an SR$\times$4 policy attains an LPIPS
of $0.243$, close to the $0.223$ of a policy trained on SR$\times$8
(Table~\ref{tab:transfer_same}). This is intuitive, since similar tasks share
inductive biases. However, they do not transfer across degradations with
different inductive biases (SR$\times$4 $\rightarrow$ HDR: $0.204$ vs.\
$0.089$; Table~\ref{tab:transfer_cross}), which would require retraining.
Instead, AHVP policies can be trained in a multi-task manner. As an
illustration, we condition the AHVP controller on a task embedding, which adds
negligible parameters, and train it jointly on SR$\times4$ and HDR. It achieves
competitive performance on both tasks compared to single-task policies
(Table~\ref{tab:transfer_cross}), showing that AHVP can amortize guidance
across multiple tasks.

\noindent
\textbf{Sensitivity to $\kappa$.}
Next, we assess the impact of the policy width $\kappa$ (Eq.~\ref{eq:gauss_policy})
on reconstruction quality. We retrain AHVP on SR$\times$4 (FFHQ) with policy width
$\kappa \in \{0.01, 0.05, 0.1, 0.2\}$ and use the same $\kappa$ at inference. We find that
the model is robust to variation in $\kappa$ during training. Similarly,
when trained with a fixed $\kappa = 0.05$, the model tolerates inference-time $\kappa$,
(Table~\ref{tab:kappa}) but excess stochasticity in the
policy starts to hurt at $\kappa = 0.2$ (LPIPS $0.091$).

\tabKappaAndTrainingCost

\noindent
\textbf{Cost of Amortization.}
Next we assess the cost of amortization. While amortization requires training the controllers, but this cost is smaller than
it may appear. First, controller training is cheap: on SR$\times$4 (FFHQ), the
controller reaches within about $5\%$ of its best LPIPS in ${\sim}4$\,h
(Table~\ref{tab:training_cost}), roughly the cost of generating (3{,}600 samples
with NDTM), and is quickly repaid whenever the sampler is reused. Second, AHVP
has few hyperparameters: the controller architecture is identical across
tasks, and only the number of DiT blocks changes between datasets (4 for FFHQ,
8 for ImageNet; Table~\ref{tab:controller_arch}), whereas training-free solvers
require per-task tuning of step sizes and guidance weights. Third, a single
task-conditioned controller trained jointly on several tasks
(Table~\ref{tab:transfer_cross}) shares the training cost across them.

\section{Related Work}
\label{sec:related}
\textbf{Test-time scaling in diffusion models} has recently gained traction in
settings ranging from text-to-image alignment
\citep{ma2025inferencetimescalingdiffusionmodels} to inverse problems
\citep{daras2024surveydiffusionmodelsinverse, zheng2025inversebench}.
Some earlier works rely on tractable approximations of the diffusion posterior
\citep{chung2022diffusion, song2022pseudoinverse, pandey2024fast,
boys2023tweedie, pokle2024trainingfree} based on Tweedie's estimate for guidance
in diffusion models. However, these approximations are often too simplistic,
require costly Jacobian–vector products, or restrict applicability to certain
downstream tasks or diffusion model architectures (latent vs. pixel space),
leading to poor sample quality, instability, or limited general use.
Consequently, some recent work \citep{Yu_2023_ICCV, bansal2024universal} adds a
correction term to these approximations to better satisfy the constraints. In
contrast, optimization-based methods \citep{he2024manifold,
rout2024rbmodulationtrainingfreepersonalizationdiffusion, pandey2025variational,
janati2024divideandconquer, zhang2025improving} embed an inner optimization loop
within the reverse diffusion process to estimate the diffusion posterior at each
timestep. However, this can be expensive during inference as it may require
backpropagation through the entire diffusion chain \citep{eyring2024reno,
dmplug}. Our method instead sidesteps such approximations and optimization loops
and allows sample generation via a single rollout of the diffusion process.

Another line of work in test-time guidance focuses on non-differentiable
rewards/constraints. For instance, SCG \citep{HuangGLHZSGOY24} leverages ideas
from path integral control for symbolic music generation. Singhal et
al.~\citep{singhal2025a} propose steering diffusion models at test-time by
sampling a collection of particles and
resampling them at intermediate steps based on their expected reward. Lastly,
Geyfman et al.~\citep{geyfman2026calibratedtesttimeguidancebayesian} propose the
REINFORCE estimator for guidance via black-box rewards for Bayesian inference.
In contrast, we focus only on differentiable terminal costs, and extending our
framework to non-differentiable costs could be an important direction for future
work.

\noindent
\textbf{Reward-based finetuning in diffusion models.} Test-time scaling adapts
the reverse diffusion process but can be computationally expensive. Inspired by
reward-based finetuning in language models \citep{ziegler2019fine,
rafailov2023direct}, recent work explores reward-based finetuning for diffusion
models \citep{black2023training, 10.5555/3618408.3618793,
uehara2024understandingreinforcementlearningbasedfinetuning}. We focus on
differentiable rewards. DRaFT \citep{clark2024directly} directly optimizes such
rewards while reducing cost through truncated backpropagation. Our method
pursues a similar objective but includes additional regularization, improving
training stability. Closely related is entropy-regularized finetuning of
diffusion models \citep{uehara2024fine, tang2024fine}. In particular, Uehara et
al.~\citep{uehara2024fine} propose a two-stage framework similar to ours with
two key differences. First, they infer the optimal initial noise using neural
SDE solvers, which can be computationally expensive at inference time, whereas
we directly predict the initial noise with a transformer. Second, our tractable
objective (Eq.~\ref{eq:elbo_final}) introduces an additional regularization
penalty on the per-step policy, leading to a different stochastic control
formulation. Building on this objective, Domingo-Enrich et
al.~\citep{domingo-enrich2025adjoint} propose a memoryless schedule for more
efficient transport between base and target distributions. DEFT
\citep{denker2024deft} instead keeps the pretrained model frozen and learns the
generalised $h$-transform with a small network. Another line of work
\citep{eyring2025noise, pmlr-v267-venkatraman25a} removes amortization in latent
space and derives a similar noise-prediction objective, though from a different
theoretical perspective. Extending such scheduling ideas to our control
formulation is an interesting future direction.

\section{Conclusion}
We introduced a principled framework for amortizing test-time guidance in
diffusion models via hierarchical variational policies. By interpreting guidance
as stochastic control over denoising trajectories, our method learns expressive
policies that steer sampling while avoiding expensive test-time optimization.
Empirically, the resulting fully amortized sampler (AHVP) achieves a strong
perceptual quality--speed tradeoff across several challenging inverse problems,
matching or exceeding recent test-time scaling baselines while requiring
substantially less computation. Our semi-amortized extension (SHVP) further
combines amortized proposals with lightweight test-time refinement, achieving
state-of-the-art perceptual quality while maintaining efficient inference. More
broadly, our results highlight amortized stochastic control as a promising
framework for efficient test-time adaptation in diffusion models. Future work
includes extending the framework to non-differentiable rewards, exploring richer
hierarchical policies, and applying amortized guidance to broader generative
tasks such as text-to-image alignment and video generation.

\begin{ack}
Stephan Mandt acknowledges funding from the National Science Foundation (NSF) through an NSF
CAREER Award IIS-2047418, IIS-2611549, the NSF LEAP Center, the Chan Zuckerberg Initiative,
SAP, and the Hasso Plattner Research Center at UCI.
\end{ack}

\bibliographystyle{plain}
\bibliography{main}

\appendix

\section{Hierarchical Variational Policies}

Here we present the proof of the bound in Eq.~\ref{eq:elbo_final} in the main
text.

\paragraph{Proof overview.}
We derive a variational lower bound (ELBO) on $\log p(\vy)$ by introducing a
structured variational distribution $q(\rvx_{0:T} \mid \vy)$ augmented with
auxiliary control variables $\{\rvu_t\}$.
The proof proceeds in three steps:
\begin{enumerate}
    \item We write the standard ELBO, decomposing it into an expected log-joint
    (energy) term and an entropy term.
    \item We lower-bound the entropy by introducing a factored approximate
    posterior $\bar{r}(\rvu_{1:T} \mid \rvx_{0:T}, \vy)$ over the controls,
    exploiting the non-negativity of the KL divergence.
    \item We substitute the factored parameterizations of both the generative
    and variational processes to arrive at a per-timestep bound.
\end{enumerate}

\begin{proof}
Recall the generative process,
\begin{align}
    p(\rvx_{0:T}, \vy) = p(\rvx_T)\left[\prod_{t=1}^{T} p(\rvx_{t-1} \mid
    \rvx_t)\right]p(\vy \mid \rvx_0).
    \label{eq:p_process}
\end{align}
We introduce a variational distribution $q(\rvx_{0:T} \mid \vy)$ and apply
Jensen's inequality to obtain the standard ELBO,
\begin{align}
    \log p(\vy)
    &= \log \int \frac{p(\rvx_{0:T}, \vy)}{q(\rvx_{0:T} \mid \vy)} q(\rvx_{0:T}
    \mid \vy)\, d\rvx_{0:T} \notag \\
    &\geq \underbrace{\E_{q(\rvx_{0:T} \mid \vy)}\left[\log p(\rvx_{0:T},
    \vy)\right]}_{\text{Joint-Likelihood}} \underbrace{- \E_{q(\rvx_{0:T} \mid
    \vy)}\left[\log q(\rvx_{0:T} \mid \vy)\right]}_{\text{Entropy}\;(\gH_q)}.
    \label{eq:proof_base_elbo}
\end{align}
We parameterize the variational distribution as a Markov chain over states:
\begin{align}
    q(\rvx_{0:T} \mid \vy) = q(\rvx_T \mid \vy)\prod_t q(\rvx_{t-1} \mid \rvx_t,
    \vy).
    \label{eq:q_markov}
\end{align}
As in the main text, we augment this variational distribution with additional
latent control variables $\{\rvu_t\}$ to make it more expressive,
\begin{align}
    q(\rvx_{0:T} \mid \vy) = \int q(\rvx_{0:T}, \rvu_{1:T} \mid \vy)\,
    d\rvu_{1:T},
    \label{eq:q_marginal}
\end{align}
where we define the augmented joint as,
\begin{align}
    q(\rvx_{0:T}, \rvu_{1:T} \mid \vy) = q(\rvx_T \mid \vy) \prod_t q(\rvu_t
    \mid \rvx_t, \rvu_{>t}, \vy) \prod_t q(\rvx_{t-1} \mid \rvx_t, \rvu_t, \vy).
    \label{eq:q_augmented}
\end{align}
Here $q(\rvu_t \mid \rvx_t, \rvu_{>t}, \vy)$ denotes the \emph{variational
policy} that selects a control conditioned on the current state $\rvx_t$, the
history of controls $\rvu_{>t}$, and the guidance signal $\vy$, while
$q(\rvx_{t-1} \mid \rvx_t, \rvu_t, \vy)$ is the controlled state transition. The
joint-likelihood term in Eq.~\eqref{eq:proof_base_elbo} maximizes the likelihood
of the generative process in Eq.~\eqref{eq:p_process}, while the entropy term
encourages exploration in the variational distribution. We simplify both terms
below, starting with the entropy.

\paragraph{Simplifying the entropy.}
We lower-bound $\gH_q$ by introducing an approximate posterior $r(\rvu_{1:T}
\mid \rvx_{0:T}, \vy)$ over the controls and exploiting $\kl{\cdot}{\cdot} \geq
0$. Let $q(\rvu_{1:T} \mid \rvx_{0:T}, \vy)$ denote the exact posterior over
controls implied by Eqs.~\eqref{eq:q_marginal}--\eqref{eq:q_augmented}. Then,
\begin{align}
    \gH_q
    &= -\E_{q(\rvx_{0:T} \mid \vy)}\!\left[\log q(\rvx_{0:T} \mid \vy)\right]
    \notag \\
    &= -\E_{q(\rvx_{0:T} \mid \vy)}\!\left[\log q(\rvx_{0:T} \mid \vy) +
    \kl{q(\rvu_{1:T} \mid \rvx_{0:T}, \vy)}{q(\rvu_{1:T} \mid \rvx_{0:T},
    \vy)}\right] \notag \\
    &\geq -\E_{q(\rvx_{0:T} \mid \vy)}\!\left[\log q(\rvx_{0:T} \mid \vy) +
    \kl{q(\rvu_{1:T} \mid \rvx_{0:T}, \vy)}{\bar{r}(\rvu_{1:T} \mid \rvx_{0:T},
    \vy)}\right],
    \label{eq:entropy_kl_bound}
\end{align}
where the bound becomes exact when $\bar{r}$ matches $q$ exactly. This can be
further simplified as,
\begin{align}
    \gH_q &\geq -\E_{q(\rvx_{0:T} \mid \vy)}\!\left[\log q(\rvx_{0:T} \mid \vy)
    + \kl{q(\rvu_{1:T} \mid \rvx_{0:T}, \vy)}{\bar{r}(\rvu_{1:T} \mid
    \rvx_{0:T}, \vy)}\right],\\
    &= -\E_{q(\rvx_{0:T} \mid \vy)}\!\left[\log q(\rvx_{0:T} \mid \vy) +
    \E_{q(\rvu_{1:T} \mid \rvx_{0:T}, \vy)}\log \frac{q(\rvu_{1:T} \mid
    \rvx_{0:T}, \vy)}{\bar{r}(\rvu_{1:T} \mid \rvx_{0:T}, \vy)}\right]\\
    &= -\E_{q(\rvx_{0:T}, \rvu_{1:T} \mid \vy)}\!\left[\log q(\rvx_{0:T} \mid
    \vy) + \log \frac{q(\rvu_{1:T} \mid \rvx_{0:T}, \vy)}{\bar{r}(\rvu_{1:T}
    \mid \rvx_{0:T}, \vy)}\right]\\
    &= -\E_{q(\rvx_{0:T}, \rvu_{1:T} \mid \vy)}\!\left[\log q(\rvx_{0:T} \mid
    \vy) + \log q(\rvu_{1:T} \mid \rvx_{0:T}, \vy) - \log \bar{r}(\rvu_{1:T}
    \mid \rvx_{0:T}, \vy)\right]
\end{align}
Expanding the KL and using $\log q(\rvx_{0:T}, \rvu_{1:T} \mid \vy) = \log
q(\rvx_{0:T} \mid \vy) + \log q(\rvu_{1:T} \mid \rvx_{0:T}, \vy) $, this
simplifies to,
\begin{align}
    \gH_q
    &\geq -\E_{q(\rvx_{0:T}, \rvu_{1:T} \mid \vy)}\!\left[\log q(\rvx_{0:T},
    \rvu_{1:T} \mid \vy) - \log \bar{r}(\rvu_{1:T} \mid \rvx_{0:T}, \vy)\right].
    \label{eq:entropy_bound_compact}
\end{align}
We factorize the approximate posterior as a product of per-timestep terms,
\begin{align}
    \bar{r}(\rvu_{1:T} \mid \rvx_{0:T}, \vy) = \prod_t \bar{r}(\rvu_t \mid
    \rvx_t, \rvu_{>t}, \vy).
    \label{eq:r_factored}
\end{align}
Substituting the parameterizations from Eqs.~\eqref{eq:q_augmented} and
\eqref{eq:r_factored} into Eq.~\eqref{eq:entropy_bound_compact}, and writing
$\E_q$ as shorthand for $\E_{q(\rvx_{0:T}, \rvu_{1:T} \mid \vy)}$, we obtain,
\begin{align}
    \gH_q
    &\geq -\E_q\!\Big[\log q(\rvx_T \mid \vy)\Big]
    - \sum_t \E_q\!\Big[\log q(\rvx_{t-1} \mid \rvx_t, \rvu_t, \vy)\Big] \notag
    \\
    &\qquad - \sum_t \E_{\rvx_t, \rvu_{>t}}\!\Big[\kl{q(\rvu_t \mid \rvx_t,
    \rvu_{>t}, \vy)}{\bar{r}(\rvu_t \mid \rvx_t, \rvu_{>t}, \vy)}\Big],
    \label{eq:simplified_entropy}
\end{align}
where $\E_{\rvx_t, \rvu_{>t}}$ denotes the expectation obtained by rolling out a
trajectory from $\rvx_T$ to $\rvx_t$, collecting all controls $\rvu_{>t}$ along
the way.

\paragraph{Simplifying the joint likelihood.}
Since $p(\rvx_{0:T}, \vy)$ does not depend on $\rvu_{1:T}$, the energy term from
Eq.~\eqref{eq:proof_base_elbo} can be equivalently written under the augmented
joint,
\begin{align}
    \E_{q(\rvx_{0:T} \mid \vy)}\!\left[\log p(\rvx_{0:T}, \vy)\right]
    &= \E_q\!\left[\log p(\vy \mid \rvx_0) + \log p(\rvx_T) + \sum_t \log
    p(\rvx_{t-1} \mid \rvx_t)\right].
    \label{eq:like_simplified}
\end{align}

\paragraph{Combining.}
Substituting Eqs.~\eqref{eq:simplified_entropy} and \eqref{eq:like_simplified}
into the ELBO (Eq.~\eqref{eq:proof_base_elbo}), we arrive at,
\begin{align}
    \gL
    &\geq \E_q\big[\log p(\vy \mid \rvx_0)\big]
    - \kl{q(\rvx_T \mid \vy)}{p(\rvx_T)} \notag \\
    &\quad - \sum_t \E_{\rvx_t, \rvu_t}\!\Big[\kl{q(\rvx_{t-1} \mid \rvx_t,
    \rvu_t, \vy)}{p(\rvx_{t-1} \mid \rvx_t)}\Big] \notag \\
    &\quad - \sum_t \E_{\rvx_t, \rvu_{>t}}\!\big[\kl{q(\rvu_t \mid \rvx_t,
    \rvu_{>t}, \vy)}{\bar{r}(\rvu_t \mid \rvx_t, \rvu_{>t}, \vy)}\big].
    \label{eq:full_bound}
\end{align}
Lastly, we model the approximate posterior over controls as independent standard
Gaussians,
\begin{align}
    \bar{r}(\rvu_t \mid \rvx_t, \rvu_{>t}, \vy) = \bar{r}(\rvu_t) = \gN(0,
    \mI_d),
    \label{eq:r_gaussian}
\end{align}
which gives the final bound,
\begin{align}
    \gL
    &\geq \E_q\big[\log p(\vy \mid \rvx_0)\big]
    - \kl{q(\rvx_T \mid \vy)}{p(\rvx_T)} \notag \\
    &\quad - \sum_t \E_{\rvx_t, \rvu_t}\!\Big[\kl{q(\rvx_{t-1} \mid \rvx_t,
    \rvu_t, \vy)}{p(\rvx_{t-1} \mid \rvx_t)}\Big] \notag \\
    &\quad - \sum_t \E_{\rvx_t, \rvu_{>t}}\!\big[\kl{q(\rvu_t \mid \rvx_t,
    \rvu_{>t}, \vy)}{r(\rvu_t)}\big].
    \label{eq:final_bound}
\end{align}
This concludes the proof.
\end{proof}

\section{Implementation Details}
\label{app:implementation}
\subsection{Reward Specification}
Here we describe the form of the likelihood $p(\vy\mid\rvx_0)$ in more detail.
For inverse problems, the likelihood can be specified as,
\begin{equation}
    p(\vy|\rvx_0) = \gN(\gA(\rvx_0), \sigma_y^2 \mI_d)
\end{equation}

\noindent
\textbf{Super-Resolution (SR)}: Following \citep{chung2022diffusion}, the
degradation operator for super-resolution can be specified as
\begin{align}
    \vy \sim \gN(\vy| \mS^{f}\rvx, \sigma_y^2 \mI),\qquad
\end{align}
where $\mS^{f}$ represents the bicubic downsampling matrix with downsampling
factor $f$. In this work, we choose $f=\{4,8\}$ for both datasets.\\

\noindent
\textbf{Random Inpainting (90\%).}
We use random inpainting with a dropout probability of 0.9 (or 90\%). For this
task, the forward model can be specified as
\begin{align}
    \vy \sim \gN(\vy| \mM\vx, \sigma_y^2 \mI_d)
\end{align}
where $\mM \in \{0, 1\}^{d \times d}$ is the masking matrix.\\

\noindent
\textbf{High Dynamic Range (HDR).} We use the HDR setup from
\citep{chung2022diffusion}. More specifically, the HDR degradation operator is
given by
\begin{equation}
    \vy = \texttt{clip}(\alpha \rvx_0 + \beta, 0, 1.0)
\end{equation}
Depending on the choice of the scalar coefficients $\alpha$ and $\beta$, the
resulting degradation can be oversaturated or undersaturated. Following prior
work \citep{zhang2025improving}, we choose $\alpha=2.0$ and $\beta=0$.

\subsection{Baseline Hyperparameters}
We optimize all baselines for the best perceptual sample quality. We use the
official code implementation from \url{https://github.com/czi-ai/oc-guidance}.
Below, we highlight the key hyperparameters for different baselines.\\

\noindent
\textbf{Note on the number of reverse diffusion sampling steps.} As mentioned in
the main text, we mainly focus on conditional sampling under low inference
budgets and therefore reduce the sampling budget of all competing baselines by a
factor of 5--6$\times$, with a minimum number of reverse diffusion steps set to
8. We elaborate more below.\\

\noindent
\textbf{DPS \citep{chung2022diffusion}.} Across all tasks and datasets, we use
DPS with the DDIM sampler and $\eta=0.5$. We reduce the number of reverse
diffusion steps in DPS from 1000 \citep{chung2022diffusion} to 200. We use the
adaptive step size proposed in \citep{chung2022diffusion} and set it to
\begin{equation}
    \zeta^{'} = \frac{\zeta}{\Vert\vy - \gA(\hat{\rvx}_0)\Vert_2^2}
\end{equation}
where we fix $\zeta=1.0$.\\

\noindent
\textbf{DDRM \citep{kawar2022denoising}.} Across all tasks and datasets,
following \citep{kawar2022denoising}, we set $\eta_b=0.85$ and $\eta=1.0$, which
we found robust across tasks. We reduce the sampling budget of DDRM from 20, as
proposed in \citep{kawar2022denoising}, to 10. DDRM can only be applied to
linear inverse problems.\\

\noindent
\textbf{C-$\Pi$GDM \citep{pandey2024fast}.} For all tasks and datasets, we fix
the number of diffusion steps to 10 using the noise conditioned C-$\Pi$GDM
sampler. Here $w$ and $\tau$ represent the hyperparameters for projection to a
conjugate space while $\tau$ represents the start-time for reverse diffusion
sampling. We find that C-$\Pi$GDM is unstable for random inpainting. A similar
observation was made by \citep{pandey2025variational}. C-$\Pi$GDM only works for
linear inverse problems.\\

\tabCPiGDMHparams

\noindent
\textbf{DAPS \citep{zhang2025improving}.} For all tasks and datasets, we use the
DAPS-200 configuration from \citep{zhang2025improving} without modification. The
number of reverse diffusion steps is set to 2, with 100 MCMC steps between each
step.\\

\noindent
\textbf{MPGD \citep{he2024manifold}.} For all tasks and datasets, we reduce the
number of steps for MPGD from 100 \citep{he2024manifold} to 20. We set DDIM
$\eta=0.5$ and the scale parameter to 20.0, which we found to work best across
tasks.\\

\noindent
\textbf{RED-Diff \citep{reddiff}.} For all tasks and datasets, we reduce the
number of steps for RED-Diff to 10. We use the DDIM sampler with $\eta=0.0$. We
highlight other important hyperparameters, such as the learning rate and
gradient-term weight, in Table~\ref{table:reddiff_hparams}.\\

\tabREDDiffHparams

\noindent
\textbf{NDTM.} We reduce the number of reverse diffusion sampling steps in NDTM
from 50 \citep{pandey2025variational} to 8. Moreover, we fix the weighting of
different loss components in NDTM to a terminal cost weight of $w_T=50$, score
regularizer, and control weights set to \texttt{ddim}. Refer to Eq. 17 in
\citep{pandey2025variational} for more details on the specific form of the loss
weighting. We tune NDTM for best performance with all task-specific
hyperparameters in Table~\ref{table:ndtm_hparams}.

\tabNDTMHparams

\noindent
\textbf{DEFT \citep{denker2024deft}.} DEFT keeps the pretrained diffusion model
frozen and learns the generalised $h$-transform with a small network whose
output is added to the pretrained noise prediction. Following the official
implementation, this network is a small U-Net (64 base channels, one residual
block per resolution, attention at resolutions 32 and 16). It takes as input the
current state $\rvx_t$, the Tweedie estimate $\hat{\rvx}_0$, and the likelihood
gradient $-\nabla_{\hat{\rvx}_0} \frac{1}{2}\Vert \vy - \gA(\hat{\rvx}_0)
\Vert_2^2$; for inpainting, it additionally receives $\vy$ and the mask. We
train it on the same images as AHVP for 2k steps at batch size 16 using Adam
with a learning rate of $5\times10^{-4}$, linearly decayed to zero, and an EMA
decay of 0.999. We sample with the EMA weights using 8 DDIM steps ($\eta=0$)
over the full diffusion schedule, starting from Gaussian noise as in
\citep{denker2024deft}.\\

\noindent
\textbf{Adjoint Matching \citep{domingo-enrich2025adjoint}.} We fine-tune the
pretrained model with Adjoint Matching by learning an additive correction to its
noise prediction, $\bm{\epsilon}(\rvx_t, t) + \bm{\delta}_\psi([\rvx_t, \gA^{+}\vy],
t)$, while keeping the pretrained network frozen. The correction network
$\bm{\delta}_\psi$ uses the same DiT architecture as our per-step controller
(Table~\ref{tab:controller_arch}) and takes the current state stacked with the
pseudoinverse $\gA^{+}\vy$ as input. The reward is $r(\rvx_0) = -w_T\Vert \vy -
\gA(\rvx_0) \Vert_2^2$. We use the Euler--Maruyama form of the Adjoint Matching
loss (App.~E.4 of \citep{domingo-enrich2025adjoint}) with memoryless DDPM
transitions ($\eta=1$) for both training and sampling. We train on the same
images as AHVP for 5k steps at batch size 16
using Adam with a learning rate of $10^{-4}$ and $\beta=(0.95, 0.999)$, gradient
clipping at norm 1.0, and an EMA decay of 0.999. We sample with the EMA weights
using 8 steps with the same initialization and timestep schedule as AHVP. For
each task, we report the better run (in LPIPS) of two terminal weights: the
$w_T$ used for AHVP, or $w_T = 1/(2\sigma_y^2) = 5000$, which recovers the
log-likelihood $\log p(\vy\mid\rvx_0)$ up to a constant. This selects
$w_T=5000$ for SR$\times$4, HDR, and SR$\times$8 on ImageNet; $w_T=50$ for
SR$\times$8 on FFHQ; and $w_T=25$ for random inpainting.

\subsection{Training Hyperparameters}
For modeling the noise predictor and the per-step controller, we use the DiT
\citep{Peebles_2023_ICCV} architecture. See Table~\ref{tab:controller_arch} for
more details. During training, we fix the learning rate to 1e-4 for both stages
(noise prediction and per-step controls) and use the Adam
\citep{kingma2017adammethodstochasticoptimization} optimizer. We highlight the
loss weights, i.e., the terminal loss ($w_T$) and the control loss
($w_\text{control}$), for different tasks for Stage-1 training in
Table~\ref{table:stage_1_hyperparams}. For Stage-2 training, we mostly adopt the
weights of terminal loss ($w_T$), control loss ($w_\text{control}$), and score
loss ($w_\text{score}$) from \citep{pandey2025variational} and specify them in
Table~\ref{table:stage_2_hyperparams}. During training, we set $\kappa=0.05$ as
the multiplicative factor for the standard deviation of the stochastic
controller (see Eq. 10 in the main text).

\tabStageOneLossWeights

\tabControllerArch

\tabStageTwoLossWeights

\section{Extended Results}
\label{app:extended_results}

\subsection{Reconstruction metrics.} We provide PSNR and SSIM comparisons
between different baselines in Table~\ref{tab:recovery}. Our method performs
comparably to most baselines across different tasks.

\subsection{Extended Qualitative Results.} We provide additional qualitative
comparisons between our proposed methods (AHVP, SHVP) and competing baselines in
Figs.~\ref{fig:app_inpainting}, \ref{fig:app_hdr}, and \ref{fig:app_sr8}.

\tabReconstructionMetrics

\figAppInpainting

\figAppHDR

\figAppSRxEight

\clearpage

\newpage
\section*{NeurIPS Paper Checklist}

\begin{enumerate}

\item {\bf Claims}
    \item[] Question: Do the main claims made in the abstract and introduction
    accurately reflect the paper's contributions and scope?
    \item[] Answer: \answerYes{}
    \item[] Justification: The claim in the abstract and introduction accurately
    reflects the contributions and scope.

\item {\bf Limitations}
    \item[] Question: Does the paper discuss the limitations of the work
    performed by the authors?
    \item[] Answer: \answerYes{}
    \item[] Justification: We discuss the limitations and future work in the
    conclusion

\item {\bf Theory assumptions and proofs}
    \item[] Question: For each theoretical result, does the paper provide the
    full set of assumptions and a complete (and correct) proof?
    \item[] Answer: \answerYes{}
    \item[] Justification: Yes, the paper provides the full set of assumptions
    and a proof in the Appendix.

    \item {\bf Experimental result reproducibility}
    \item[] Question: Does the paper fully disclose all the information needed
    to reproduce the main experimental results of the paper to the extent that
    it affects the main claims and/or conclusions of the paper (regardless of
    whether the code and data are provided or not)?
    \item[] Answer: \answerYes{}
    \item[] Justification: All the details for training and inference are
    discussed in the Appendix and the main text.

\item {\bf Open access to data and code}
    \item[] Question: Does the paper provide open access to the data and code,
    with sufficient instructions to faithfully reproduce the main experimental
    results, as described in supplemental material?
    \item[] Answer: \answerYes{}
    \item[] Justification: We will provide access to our code upon acceptance.

\item {\bf Experimental setting/details}
    \item[] Question: Does the paper specify all the training and test details
    (e.g., data splits, hyperparameters, how they were chosen, type of
    optimizer) necessary to understand the results?
    \item[] Answer: \answerYes{}
    \item[] Justification: We specify all these details in the Appendix.

\item {\bf Experiment statistical significance}
    \item[] Question: Does the paper report error bars suitably and correctly
    defined or other appropriate information about the statistical significance
    of the experiments?
    \item[] Answer: \answerNo{}
    \item[] Justification: We do not report error bars as different runs are
    expensive to repeat

\item {\bf Experiments compute resources}
    \item[] Question: For each experiment, does the paper provide sufficient
    information on the computer resources (type of compute workers, memory, time
    of execution) needed to reproduce the experiments?
    \item[] Answer: \answerYes{}
    \item[] Justification: We provide all the details in the Appendix.

\item {\bf Code of ethics}
    \item[] Question: Does the research conducted in the paper conform, in every
    respect, with the NeurIPS Code of Ethics
    \url{https://neurips.cc/public/EthicsGuidelines}?
    \item[] Answer: \answerYes{}
    \item[] Justification: Yes we conform with the NeurIPS code of Ethics.

\item {\bf Broader impacts}
    \item[] Question: Does the paper discuss both potential positive societal
    impacts and negative societal impacts of the work performed?
    \item[] Answer: \answerNA{}
    \item[] Justification: There is no societal impact of the work performed

\item {\bf Safeguards}
    \item[] Question: Does the paper describe safeguards that have been put in
    place for responsible release of data or models that have a high risk for
    misuse (e.g., pre-trained language models, image generators, or scraped
    datasets)?
    \item[] Answer: \answerNA{}
    \item[] Justification: The paper poses no such risks.

\item {\bf Licenses for existing assets}
    \item[] Question: Are the creators or original owners of assets (e.g., code,
    data, models), used in the paper, properly credited and are the license and
    terms of use explicitly mentioned and properly respected?
    \item[] Answer: \answerYes{}
    \item[] Justification: We acknowledge the code resources used wherever
    needed

\item {\bf New assets}
    \item[] Question: Are new assets introduced in the paper well documented and
    is the documentation provided alongside the assets?
    \item[] Answer: \answerNA{}
    \item[] Justification: We do not release any new assets.

\item {\bf Crowdsourcing and research with human subjects}
    \item[] Question: For crowdsourcing experiments and research with human
    subjects, does the paper include the full text of instructions given to
    participants and screenshots, if applicable, as well as details about
    compensation (if any)?
    \item[] Answer: \answerNA{}
    \item[] Justification: We do not involve crowdsourcing or working with human
    subjects.

\item {\bf Institutional review board (IRB) approvals or equivalent for research
with human subjects}
    \item[] Question: Does the paper describe potential risks incurred by study
    participants, whether such risks were disclosed to the subjects, and whether
    Institutional Review Board (IRB) approvals (or an equivalent approval/review
    based on the requirements of your country or institution) were obtained?
    \item[] Answer: \answerNA{}
    \item[] Justification: The paper does not involve crowdsourcing nor research
    with human subjects.

\item {\bf Declaration of LLM usage}
    \item[] Question: Does the paper describe the usage of LLMs if it is an
    important, original, or non-standard component of the core methods in this
    research? Note that if the LLM is used only for writing, editing, or
    formatting purposes and does \emph{not} impact the core methodology,
    scientific rigor, or originality of the research, declaration is not
    required.
    \item[] Answer: \answerNA{}
    \item[] Justification: The paper did not use any LLMs for core
    methodological development.

\end{enumerate}

\end{document}